# Binary stochasticity enabled highly efficient neuromorphic deep learning achieves better-than-software accuracy

Yang Li[1], Wei Wang[1*], Ming Wang[2], Chunmeng Dou[3], Zhengyu Ma[1], Huihui Zhou[1], Peng Zhang[1], Nicola Lepri[4], Xumeng Zhang[2], Qing Luo[3], Xiaoxin Xu[3], Guanhua Yang[3], Feng Zhang[3], Ling Li[3], Daniele Ielmini[4], and Ming Liu[2,3]

[1]Peng Cheng Laboratory, Shenzhen 518000, China;

[2]Frontier Institute of Chip and System, State Key Laboratory of Integrated Chips and Systems, Zhangjiang Fudan International Innovation Center, Fudan University, Shanghai 200433, China;

[3]Institute of Microelectronics, Chinese Academy of Sciences, Beijing 100029, China;

[4]Dipartimento di Elettronica, Informazione e Bioingegneria, Politecnico di Milano, Milano 20133, Italy.

*Email: wangwei@pcl.ac.cn

**Abstract**

Deep learning[1–3] needs high-precision handling of forwarding signals, backpropagating errors, and updating weights. This is inherently required by the learning algorithm since the gradient descent learning rule relies on the chain product of partial derivatives[4,5]. However, it is challenging to implement deep learning in hardware systems that use noisy analog memristors as artificial synapses, as well as not being biologically plausible[6–9]. Memristor-based implementations generally result in an excessive cost of neuronal circuits[10] and stringent demands for idealized synaptic devices[11–14]. Here, we demonstrate that the requirement for high precision is not necessary and that more efficient deep learning can be achieved when this requirement is lifted. We propose a binary stochastic learning algorithm that modifies all elementary neural network operations, by introducing (i) stochastic binarization of both the forwarding signals and the activation function derivatives, (ii) signed binarization of the backpropagating errors, and (iii) step-wised weight updates. Through an extensive hybrid approach of software simulation and hardware experiments, we find that binary stochastic deep learning systems can provide better performance than the software-based benchmarks using the high-precision learning algorithm. Also, the binary stochastic algorithm strongly simplifies the neural network operations in hardware, resulting in an improvement of the energy efficiency for the multiply-and-accumulate operations by more than three orders of magnitudes.

**Main**

By adopting the in-memory computing paradigm, a neuromorphic system based on synaptic weights of analog memristors[15] can parallelly and efficiently accelerate the massive high-precision multiply-and-accumulate (MAC) operations[16] required in deep learning. The MAC operations of each neural network layer in the forward or backward propagations can be parallelly accomplished in one step based on Kirchhoff's current law and Ohm's law[15,17], where the input signals are represented as voltages, the weights between the input nodes and output nodes are stored as conductance in a crossbar array of memristors and the output signals are represented as currents. However, to complete this task, one needs to accurately tune the input voltages and accurately sense the output currents, which require high-precision digital-to-analog and analog-to-digital converters, respectively. This results in high power consumption and a large circuit footprint, thus counterbalancing the computation acceleration gained by the parallel in-memory and analog computation[10]. Besides, while gradient calculation and weight update in each neural network layer can be executed in parallel[18,19], the precise, gradual tuning of conductance of the memristor synapses is challenging due to device variations. Thus, efficient online, in-memory learning in a stand-alone non-von Neumann architecture becomes a prohibitive task. Moreover, in a biological system, neural signals are transmitted as spikes, i.e., all-or-none action potentials[6,7], which mathematically translates in a binary format of signal amplitude. Finally, the synaptic weights are implemented in analog and noisy components[8,9], subject to stochastic fluctuations.

In contemporary deep learning theory, the high-precision handling of the signals and errors is an inherent requirement since the gradient-descent learning rule relies on the product of a partial derivative chain, i.e., the chain rule of calculus[4,5]. High-precision numbers in the computing system are needed to accurately represent the continuous changes of the signals and errors, while high-precision synaptic weights are needed to accurately accumulate the descending gradient. Specialized algorithms such as neural network quantization[20] and binarization[21] mainly aim at reducing the computational cost in the inference stage. In the learning stage, estimation of partial derivatives and high-precision description of error signals are essential[22,23]. Similarly, to train a deep spiking neural network, the surrogate gradient learning method needs a surrogate representation of derivatives and errors with sufficient accuracy[24].

In this article, we go beyond the high-precision requirement of deep learning and propose a binary stochasticity-based hardware-friendly approach. Firstly, we stochastically binarize both the forwarding signals and the derivatives of the activation function in each layer. Instead of viewing the stochastic binarization as a non-differentiable function and attempting to estimate its derivative[22], we view the stochastic binarization of the signal as equivalent to its floating-point representation. Secondly, only the sign of the errors are backpropagated while any error magnitude information is ignored, thus enabling the highly efficient hardware implementation of the error backpropagation. Finally, we propose a periodical-carry step-wise weight update method, supporting the in-memory deep learning using noisy and fluctuating memristor synapses. When implementing this algorithm in a crossbar array of analog memristors, with the help of external or intrinsic noise, stochastic binarization can be accomplished by using highly simplified peripheral/neuronal circuits. The calculations of the activation function and its derivative are not needed, nor is the explicit and accurate sensing of the outputs (i.e., the electrical currents) of the crossbar array. Thus, the complex and expensive analog-to-digital and digital-to-analog converters are eliminated from the neuronal circuits. The step-wise weight update method inherently enables the quantization of the weight during the training, which guarantees a large tolerance to the noisy and non-linear synaptic plasticity of analog memristors. We systematically investigate the effect of these algorithms in fully-connected and convolutional deep neural networks for MNIST and CIFAR-10 datasets. When implementing the proposed binary stochastic algorithms in an analog memristor-based neuromorphic system, the energy efficiency of deep learning tasks can be improved by two orders of magnitudes compared to traditional memristive neuromorphic algorithms. Compared to traditional high-precision algorithms using CMOS technology in the graphical processing units of von-Neumann architecture, using the proposed algorithms, an analog memristor-based neuromorphic system improves energy efficiency by more than three orders of magnitudes. The high efficienct neuromorphic deep learning system unexpectively achieves better-than-software accuracy.

**Hardware-friendly and biologically plausible algorithms for deep learning**

Deep neural networks compute via multiple layers of neurons interconnected by tunable weights. Signals and errors propagate through the layers in the forward and backward directions, respectively, while the learning is achieved by updating the weights of each layer via the gradient descent rule. Our neural networks rely on three key concepts, namely i) stochastic binarization of the forward propagating signals (**Figure 1a**); ii) stochastic binarization of the activation derivatives (Figure 1b); and iii) signed binarization of the backpropagating errors (Figure 1c).

**Stochastic binarization of the forwarding signals.** In the forward pass, signals of a training sample are transmitted layer-by-layer to obtain a tentative output in the last layer. Within a typical layer $l$, the state of a neuron $j$ is a non-linear function of the weighted summation ($y_j^l$, i.e., the membrane potential) of input signals ($x_i^l$) from the previous layer, denoted by $z_j^l = \sigma(y_j^l) =$ $\sigma(\sum_i x_i^l w_{i,j}^l)$, where $w_{i,j}^l$ is the transmitting weight (synapse) from the $i$th input signal to the neuron $j$, and $\sigma(\cdot)$ is the activation function (Figure 1a). Instead of directly transmitting the $l$th layer's output as the ($l$+1)th layer's input, that is $x_j^{l+1} = z_j^l$, we activate the intra-layer transmitting signal by a stochastic binarization process (i.e., the Bernoulli process), where the transmitting signal is activated (state "1") with a probability of $z_j^l$ and is deactivated (state "0") otherwise, that is $P(x_j^{l+1} = 1) = z_j^l$ (Figure 1a). Note that the input signals ($x_i^l$) should have been in binary states since the previous layer follows the same stochastic binarization rule.

The layer-wise forward propagation can be mapped to a crossbar array of memristors, where the input signals are denoted by the voltages ($V_i$) applied on the top parallel electrodes, the transmitting weights correspond to the conductance ($G_{ij}$) of the memristors in the intersections of top and bottom parallel electrodes, and the currents ($I_j$) from the bottom electrodes represent the membrane potential (Figure 1d)[17,25,26]. The crossbar array inherently performs weighted summation $I_j = \sum_i V_i G_{ij}$, where the multiplications and summations are governed by Ohm's law and Kirchhoff's current law, respectively. In the conventional feedforward propagation, the input voltages ($V_i$) are accurately tuned according to the high-precision input signals. The output currents $I_j$ should be sensed with sufficient accuracy to allow the processing of the membrane potential by the activation function. In our algorithm, instead, the input voltage is binarized with only two values, namely 0 V and a fixed read voltage $V_0$. The stochastically binarized output signals ($x_j^{l+1}$) are directly obtained by comparing the output currents $I_j$ of the crossbar array with a noise current signal $I_{noise}$ (see Figure 1d, Methods and Extended Data Figure 1a and Extended Data Figure 2a-2c).[27,28] It should be noted that the stochastic binarization of $I_j$ is consistent with the biological neuron in the human brain, where activation occurs when the membrane potential reaches a noisy threshold. The activation function provides the probability of a neuron to be activated[29], thus can be viewed as the cumulative density function of the stochastic threshold of a McCulloch-Pitts neuron[30,31] (Methods and Extended Data Figure 2c). In this sense, the activation function should always be a monotonically non-decreasing function bounded between 0 and 1.

**Stochastic binarization of the activation derivatives.** The activation function's derivative $\frac{dz_j^l}{dy_j^l} = \sigma'(y_j^l)$, which is required to complete the backward propagation, is stochastically binarized in our algorithm via a Bernoulli process. In particular, for a logistic activation function $z_j^l = \frac{1}{1+\exp(-y_j^l)}$, its derivative has a simple form of $\frac{dz_j^l}{dy_j^l} = z_j^l(1 - z_j^l)$. Instead of directly using the activation derivative, we binarize it into "1" with a probability of $z_j^l(1 - z_j^l)$ and "0" otherwise (Figure 1b).

To map this binarized backpropagation in hardware (Figure 1e), we use two consecutive flip-flops to independently sample the stochastically binarized transmitting signal ($x_j^{l+1}$) and process the sampled binary results of the flip-flops by a logic gate. The derivative $\frac{dz_j^l}{dy_j^l}$ takes the value of "1" only when the first flip-flop is "1" and the second flip-flop is "0". The probability of the derivative $\frac{dz_j^l}{dy_j^l}$ being "1" is the product of the probability of the first flip-flop being "1" ($z_j^l$) and the probability of the second flip-flop being "0" ($1 - z_j^l$), which meets the algorithm's requirement exactly (see Methods and Extended Data Figure 1b and Extended Data Figure 2d).

**Signed binarization of the backpropagating errors.** In the backward pass, the errors between the tentative output and target output in the last layer backpropagate all the way back to the first layer. As shown in Figure 1c, in our algorithm, only the signs of the post-layer errors $\delta x_j^{l+1}$ are transmitted to the neurons of the current layer, i.e., $\delta z_j^l = sign(\delta x_j^{l+1})$ where $\delta z_j^l$ is equal to 1 when $\delta x_j^{l+1}$ is nonnegative, otherwise it is equal to -1. According to the chain rule of partial derivatives, within the layer $l$, the errors of membrane potentials $\delta y_j^l$ are the product of the neuron errors $\delta z_j^l$ and the activation function's derivative $\frac{dz_j^l}{dy_j^l}$, i.e., $\delta y_j^l = \delta z_j^l \frac{dz_j^l}{dy_j^l}$, and the errors of input signals $\delta x_i^l$ are the weighted summation of the errors of membrane potential, i.e., $\delta x_i^l = \sum_j \delta y_j^l w_{i,j}^l$ (Figure 1c). Note that the errors of membrane potentials $\delta y_j^l$ are ternary valued ("-1", "0" or "1"). Note that $\delta x_j^{l+1}$ stands for $\frac{\partial L}{\partial x_j^{l+1}}$ for simplicity, where $L$ is the cross entropy loss between the actual output and the target output of a training sample. The same nomenclature applies to other variables, such as $\delta y_j^l$, $\delta z_j^l$ and $\delta x_j^l$.

When implemented in hardware, the voltages ($V_j^b$) representing the errors of the membrane potential ($\delta y_j^l$) are applied to the bottom electrodes, while the currents $I_i^b$ are collected at the top electrodes. In the traditional algorithm where the errors from the post-layer are directly transmitted to the current layer, i.e., $\delta z_j^l = \delta x_j^{l+1}$, and the derivatives are in the original form, i.e., $\frac{dz_j^l}{dy_j^l} = z_j^l(1 - z_j^l)$, the errors of membrane potential $\delta y_j^l$ should be calculated in the digital or analog domain with sufficient accuracy. Thanks to the binarized activation derivatives and the signed errors, our algorithm results in a strong simplification of the peripheral circuits for the error backpropagation (Figure 1f and Extended Data Figure 1c). The sign operation can be performed by comparing the output current ($I_j^b$) of the crossbar array with a zero current, to yield a negative or positive output voltage representing the signed error $\delta z_j^l$. A single transistor can carry out the multiplication between the signed errors and the binary derivatives. The ternary errors of membrane potentials are then represented in the states of the negative voltage ("-1"), the high impedance ("0"), and the positive voltage ("1"). The ternary output is then converted to a voltage $V_j^b$ with amplitudes $-V_0, 0$, and $V_0$ to be applied to the bottom electrodes of the crossbar array.

Weight updates are performed after the completion of the forwarding and backpropagation passes for a batch of training samples, according to the gradient descent rule $w_{i,j}^l \leftarrow w_{i,j}^l - \eta \langle \delta w_{i,j}^l \rangle_{batch}$, where $\eta$ is the learning rate, $\delta w_{i,j}^l = \frac{\partial L}{\partial w_{i,j}^l} = x_i^l \frac{\partial z_j^l}{\partial y_j^l} \delta z_j^l$ is the partial derivative of the loss function $L$ to the weight and $\langle \delta w_{i,j}^l \rangle_{batch}$ is the average $\delta w_{i,j}^l$ over a training batch.

**Binary stochasticity improves the learning performance**

We trained a fully-connected three-layered neural network for the classification of the handwritten digits from the modified National Institute of Standard Technology (MNIST[5]) dataset (**Figure 2a** and Methods). The same network was trained with either high precision (HP) or binary stochastic (BS) approaches. In high-precision learning, all signals, derivatives, and errors were represented with 32-bit floating point (FP) numbers, whereas in binary stochastic learning the forwarding signals (0 or 1), activations derivatives (0 or 1), and backpropagating errors (-1 or 1) were all mapped with binary states according to the algorithms in Figure 1a-c. More details about the high-precision and binary stochastic learning algorithm are given in Extended Data Table 1. For high-precision learning, the weight update rule can be written as

$$w^l \leftarrow w^l - \eta \underbrace{x^l}_{FP} \underbrace{\frac{\partial z^l}{\partial y^l}}_{FP} \underbrace{\delta z^l}_{FP}, \qquad (1)$$

whereas, in binary stochastic learning, the weight update follows

$$w^l \leftarrow w^l - \eta \langle \underbrace{x^l}_{\{0,1\}} \underbrace{\frac{\partial z^l}{\partial y^l}}_{\{0,1\}} \underbrace{\delta z^l}_{\{-1,1\}} \rangle_{batch} \qquad (2)$$

where the subscripts are disregarded for legibility. According to the law of total expectation, the two weight update rules are equivalent: in fact, the three terms represented in floating point precision in Eq. (1) are expectations of the three binarized terms in Eq. (2).

Binary stochastic learning achieves better performance compared to high-precision learning (Figure 2b and 2c). Although the cross entropies between the final output and the target output in the binary stochastic learning are higher than that in the high-precision learning, they both monotonically decrease as a function of the training epoch, thus showing a good learning convergence (Figure 2b). The high precision learning shows an obvious over-training effect than the binary stochastic learning since the test error on the hand-written digits from the training set quickly diminishes to 0. However, the test error on unseen hand-written digits from the test set is lower for BS learning, thus highlighting the higher generalization capability of the binary stochastic approach(Figure 2c).

The binary stochastic learning can adjust the synapse weights in earlier layers more efficiently (Figure 2d and 2e, Extended Data Figure 3a, 3d, and 3e). This can be explained by the signed operations normalizing the backpropagating errors thus preventing the error vanishing issue for the earlier layers. In high-precision learning, the activations of neurons in each layer $z^l$ tend to segregate near 0 or 1 as the learning proceeds (Figure 2f). This tendency is more obvious in binary stochastic learning, which highlights that the stochastic binarization well preserves the information of forwarding signals (Figure 2f and 2g, Extended Data Figure 3b and 3c).

To check the individual effect of the binary stochastic algorithms, we permutationally combine them with the high-precision ones and tested the learning performance of these partially binary stochastic trained networks (Extended Data Figure 4). The results show that the binary stochasticity in signal forwarding can prevent overtraining and improve the test accuracy of the network on unseen data (Extended Data Figure 4a-4c). The signed binarization of backpropagation errors is instead responsible for effectively propagating the errors to earlier layers of the network (Extended Data 4d-4k). The stochastic binarization of the activation function derivative does not show a significant effect on the learning process. We also tested binary stochastic learning algorithms using various activation functions and derivative functions (Extended Data Figure 5). The results show that, as long as the activation has a sigmoidal shape and the derivative function has a bell shape, the binary stochastic learning algorithm converges to high accuracy (Extended Data Figure 5 and Discussion).

**Binary stochasticity improves inference performance**

To assess the impact of stochastic binarization on inference, we compared three inference methods, namely (i) one-time inference using high-precision forwarding signals (HP inference, **Figure 3a**), (ii) one-time inference using deterministic binarized forwarding signals (binary inference, Figure 3b), and (iii) majority voting of repeated inferences using binary stochastic forwarding signals (stochastic inference, Figure 3c). In the stochastic inference method, results of repeated inference using binary stochastic forwarding signals are obtained and a final recognition decision is made by voting. Only the unseen data are used to test the inference performance.

In the first neural network which is trained by high-precision learning (Figure 3d), the high-precision inference has the lowest inference error. The binary inference has a higher inference error but largely simplifies neural operations and hardware circuits. The stochastic inference has the highest one-time inference error, although the inference error can be dramatically decreased by repeating the stochastic inference and taking the majority vote as the final recognition result. Overall, the accuracy of the stochastic inference can be much lower than the binarized inference and approximate the one of high precision inference.

In the second neural network which is trained by binary stochastic learning (Figure 3e), all three inference methods have lower inference errors compared to each method in the first neural network, respectively. Surprisingly, the high-precision inference test error is no longer the asymptotic line for the stochastic majority vote inference method: the stochastic inference achieves lower inference error than high-precision inference after 15 times votes (Figure 3e).

Compared to the error of the high-precision inference in the first neural network, which is about 1.57%, the stochastic inference in the second neural network has the lowest inference error (1.21%) (Figure 3f). A better performance is achieved in multiple learning schemes that permutate the information representation methods as shown in Extended Data Figure 6.

**Binary stochasticity is efficient in deep convolutional neural networks**

To study the efficiency of binary stochastic learning in convolutional neural networks, we considered the network shown in **Figure 4a**, consisting of two convolutional layers, two max-pooling layers, and one fully-connected layer, for learning and recognizing the handwritten digits from the MNIST dataset. The binary stochastic learning performance, which is shown as the test error on the test set during the training, is better than the one of the high-precision learning (Figure 4b), similar to the fully-connected neural network on the same data set. For the neural network trained with a high-precision learning algorithm, the high-precision inference shows the best performance and is taken as the baseline (Figure 4c). For the neural network trained with a binary stochastic learning algorithm, the inference performance in all cases has been improved dramatically and the performance of the stochastic inference exceeds the one of the high-precision learning after 10 times repetition of the inferences (Figure 4d). Figure 4e shows the summary of the various inference results for the neural network trained with the two learning algorithms. Over-training is completely avoided, and more salient features are learned in the convolutional kernels when using binary stochastic learning (Extended Data Figure 7).

Figure 4f shows a deeper convolutional neural network in VGG style[32] consisting of six convolutional layers, three max-pooling layers, and three fully-connected layers, for learning and recognizing the images from the CIFAR-10 dataset[33]. The high-precision learning shows better learning performance than binary stochastic learning (Figure 4g). For the neural network trained with a high-precision learning algorithm, while the high-precision inference shows the best performance (the baseline), the binary inference and the stochastic inference are no longer suitable (Figure 4h). For the neural network trained with a binary stochastic learning algorithm, while the high-precision inference and the binary inference show similar accuracies, the stochastic inference quickly exceeds them after several repetitions (Figure 4i). Figure 4j shows the summary of the various inference results for the neural network trained with the two learning algorithms.

Note that, for a fair comparison, no other learning performance enhancement techniques, such as dropout, batch normalization, or data preprocessing methods, are employed in the deep neural network for both learning algorithms.

**Quantized weights and analog weights using memristors**

Thanks to binarization operations in signal forwarding, activation derivative, and error backpropagation, the partial derivative of the loss function $L$ of a single training sample to the weight, $\delta w_{i,j}^l$, that is, the gradient of the loss to the weight, $\partial L/\partial w_{i,j}^l$, has a ternary value. However, to stabilize the learning procedure, the gradient is averaged over a batch of training samples, according to Eq. (2). Thus, the weight needs to be updated or tuned with sufficient precision. A step-wise update of the weight is generally beneficial since it is compatible with quantized weights (e.g., integers) or noisy and analog weights (e.g., analog memristors).

To achieve this goal, we used a periodical carry method[11,34] to update the weight, as illustrated in **Figure 5a**. The gradient of the loss to the weight, $\partial L/\partial w_{i,j}^l$, is accumulated in a digital counter. The weight is updated with a fixed step when and only when the accumulated gradient reaches a positive threshold ($th_+$) or a negative threshold ($th_-$). For instance, the weight can be represented by an 8-bit signed integer (INT8, taking a value from -128 to 127). When the accumulated gradient reaches the positive threshold, the corresponding weight subtracts 1, and *vice versa*. The accumulated gradient is cleared whenever such a weight update event happens. The learning rate is thus defined by the thresholds and a scaling factor between the integer weight and the effective weight (Methods).

Using the analog memristor as synaptic weight, the weight update can be largely simplified compared to the conventional iterative write and verify which is usually employed to tune the conductance of the memristor in sufficient accuracy[10,35]. In our software-based in-situ deep learning experiments, the long-term potentiation (LTP) and long-term depression (LTD) behaviors of a real memristor under identical pulses[36] were used to verify the deep learning performance. The memristor device, like the biological synapses, has high fluctuations and nonlinear plasticity under identical potentiation pulses (Figure 5b left side) and identical depression pulses (Figure 5b right side). Using the periodical carry method[13,14], when the accumulated gradient reaches the positive threshold, a depression pulse is applied to the corresponding memristor in the crossbar array, and *vice versa*. The conductance of the memristor is updated as is, or in other words, blindly, regardless of the nonlinearity and fluctuation of the weight changes: we do not read the initial and the updated conductance to verify the correctness of the amplitude and direction of the weight changes.

Figure 5c shows the learning performance of a fully-connected neural network (same as in Figure 2a) for binary stochastic learning using weights of 32-bit floating point (FP32) numbers (without the periodical carry), 8-bit signed integers (INT8), 4-bit signed integers (INT4), ternary values (-1, 0, 1), and analog memristors, compared with the high-precision learning (the baseline). More details about the neural network setups and the training results are reported in the Methods and Extended Figure 8. The analog memristor-based neural network shows an inference accuracy (98.82%) higher than the baseline (98.43%) and slightly lower than the binary stochastically learned neural network based on floating-point weights (98.88%), as summarized in Figure 5d. It should be emphasized that the neural network is trained in situ in artificial synapses of analog memristors which have noisy plasticity just like their biological counterparts.

**Discussion**

Introducing stochasticity to a neural network has been proven to be beneficial in several aspects, for instance, escaping from local minima in the Boltzmann machine[37] and preventing the overfitting effect in the dropout technique[38]. Stochasticity has a similar role in this work, resulting in better learning performance. The better performance by the majority vote of repeated stochastic inference than the high-precision inference (Figure 3e, Figure 4d, and Figure 4i), requires deeper insight. One might expect that the high-precision inference should be the asymptotic line of the repeated stochastic inference, in the sense that a real number between 0 and 1 in sufficient precision contains all the information of a neuron state. The sampled binary-valued numbers, which take the high-precision number as their probability, can restore the full information of the high-precision number only after sufficient times of repetition. However, the neural network experiments show that a group of sampled binary-valued numbers after only a few repetitions could already convey more information than the real numbers.

Binarization or, more generally, quantization of the forwarding signals has been extensively investigated to reduce the computational loads for inference in edge applications[20]. However, the lower computational demand generally comes at the cost of a lower inference accuracy. Neural networks trained with binarized or quantized activation, i.e., quantization-aware training, need to estimate or surrogate the derivative of the non-differentiable activation functions[23,24]. Here, we use the stochastically binarized state of the neurons as another representation of the activation value, with no need to estimate any derivative. In other words, the "straight-through estimator"[22] should be a straightforward solution to the non-differential activation function issue, providing that the binarization is stochastic and the activation function is sigmoid and bounded between 0 and 1. The correctness of the neural network learning is guaranteed by the law of total expectation according to which Eqs. (1) and (2) should be equivalent.

Reduction of the computation demand and the energy cost can be achieved in multiple aspects according to the specific application scenarios of binary stochastic learning. For instance, if implemented in CMOS technology[39] and using the weights represented by 32-bit floating point numbers, the binary stochastic learning algorithm reduces the energy consumption of the elementary multiply and accumulate (MAC) operation from 4.6 pJ to 0.9 pJ (Extended Data Table II). By quantizing the weight, the energy consumption could be reduced further to 0.03 pJ, 15 fJ, and 5.6 fJ for weights implemented in INT8, INT4, and ternary values, respectively (Methods). INT4 weights allow reducing the energy consumption by 307 times without sacrificing the learning performance, while ternary-valued weights reduce the energy consumption by 821 times with only a slight degradation of learning accuracy. When implementing the binary stochastic learning in a memristor-based neuromorphic system, the energy consumption for a single MAC is reduced from 0.18 pJ to 1.8 fJ (100 times reduction) since the input is in 1-bit binary state and the production does not need analog to digital converters (ADCs)[10] (Methods). Reduction of the footprint of the circuits in integrated chips by 57.8 times could also be projected. Similarly, the reduction of the footprint in similar orders of magnitudes can be achieved. Note that, the neural network implementation in CMOS technology needs to retrieve the weight data from memory frequently, whereas the weight data are stored in the memristor devices in the memristor-based neuromorphic system. Thus, conservatively, more than three orders of magnitudes of energy reduction can be obtained using the proposed binary stochastic learning algorithm combined with the memristive neuromorphic technology. Better learning performance is achieved alongside the reduction of energy consumption and footprints.

## Conclusion

In summary, we have shown that deep learning algorithms can be reformulated to be more biologically plausible and hardware friendly for neuromorphic implementation. We stochastically binarize the forwarding signals and the activation function derivatives. Only the signs of the errors are backpropagated in the backward pass. This algorithm largely simplifies the neural network operations and results in higher deep learning accuracy. Additionally, the stochastic binarization in the forwarding pass also results in better inference accuracy. Mathematically, we prove the correctness of the learning algorithm by the law of total expectation, avoiding the derivative estimation of the non-differential activation functions. We also provide a new view that the activation function should be understood as the probability of the neurons being activated, and the stochastically activated neurons in binary states are more informative than the activation function in real numbers. Finally, a periodical carry strategy is employed to quantize the weight during the learning and adapt the deep learning algorithm to be tolerant to the fluctuation and noisy synapses based on analog memristors. The energy efficiency for deep learning tasks is improved by more than three orders of magnitudes, in addition to better deep learning accuracy.

## Data availability

The datatsets for training and testing the neural networks in this study, MNIST[5] and CIFAR10[33], are publicly available. No additional datasets were generated during the current study. The data used to produce the plots within this paper are available as the Source Data upon acceptance of this paper. All other data used in this study are available from the corresponding author upon reasonable request.

## Code availability

The codes used in this study will be available in a public repository upon acceptance of this paper.

**Acknowledgments**

The authors thank for the funding from the basic and frontier research project of the Peng Cheng Laboratory, the major key project of the Peng Cheng Laboratory PCL2021A08, and the major key project of the Peng Cheng Laboratory PCL2021A13. Z.M. thanks the National Natural Science Foundation of China (NSFC) under grant 62206141. The project is also partially supported by the National Key R&D Program of China under Grant (2021YFB3601200) and the National Nature Science Foundation of China (62104042).

**Author information**

Contributions: W.W. conceived the concept of this work. Y.L. and W.W. conducted the software experiments. W.W. developed the synaptic behavior model of memristors. W.W. and Y.L. wrote the first draft of the manuscript. Z.M. and H.Z. contribute to the biologically plausible explanations of the learning algorithms. All authors discussed the results and contributed to the preparation of the manuscript.

Corresponding author: Wei Wang.

**Ethics declarations**

Competing interests: The authors declare no competing interests.

**Additional information**

None.

## METHODS

### Crossbar array of memristors for signed weight matrix

Assuming that a typical fully-connected layer (labeled as $l$) has $n$ neurons that process forwarding information from $m$ neurons of the previous layer (layer $l-1$), in memristor-based hardware implementation, the vector-matrix multiplications in both forward pass and backward pass can be implemented by a crossbar array of memristors which has $m+1$ horizontal bars/wires in the top and $n+1$ vertical bars/wires in the bottom. In each intersection of the horizontal bars and vertical bars except the intersection of the $(m+1)$th horizontal bar and the $(n+1)$th vertical bar, there is one memristor (or programmable resistor). The conductances of the memristors in the intersections of the 1-to-$m$ horizontal bars and the 1-to-$n$ vertical bars are denoted as $G_{i,j}$, where $i=1,\dots,m$ and $j=1,\dots,n$. The memristors in the intersections of the 1-to-$m$ horizontal bars and the $(n+1)$th vertical bar and the intersections of the $(m+1)$th horizontal bar and the 1-to-$n$ vertical bars are called reference memristors, which have fixed conductance denoted as $G_{ref}$ (Extended Data Figure 1a and 1c). The reference memristors are suppressed in the illustration of Figure 1d and 1f and the discussion in the main text for legibility. They are needed to form differential pairs since the synaptic weights are signed numbers while the device conductance is always positively valued. The synapse weights $w_{i,j}^l$ are represented by the conductance difference between regular memristors and reference memristors $G_{i,j} - G_{ref} = G_0 w_{i,j}^l$, where $G_0$ is a scaling factor. Assuming that the memristor has the maximal conductance and minimal conductance of $G_{max}$ and $G_{min}$, respectively, the reference conductance is set to the middle point of the memristor conductance, i.e., $G_{ref} = \frac{G_{max}+G_{min}}{2}$. To represent the weight values in the range between $w_{min}$ and $w_{max}$, the scaling factor $G_0$ is defined as $\frac{G_{max}-G_{min}}{w_{max}-w_{min}}$. $w_{min}$ and $w_{max}$ are empirical parameters and are takes the default values of -1 and 1, respectively.

### Forward weighted summation by the memristor crossbar

If the previous layer has already binarized the forwarding signals, that is, the input information $x_i^l$ in this layer is in binary states, the weight summation for the membrane potential $y_j^l = \sum_i^m x_i^l w_{i,j}^l$ could be easily implemented. The binary states, "1" or "0", are represented by logic voltage levels with an amplitude of $V_{DD}$ or 0V, respectively, where $V_{DD}$ is the supply voltage of the digital circuits. Since $V_{DD}$ is too large to be directly applied to the memristor array without changing the conductance of memristors, level shifters are used to convert $V_{DD}$ to a read voltage with the amplitude of $V_0$. Thus, the voltages $V_i = x_i^l V_0$ ($i = 1, \ldots m$) are applied to the $i$th top bar of the memristor array (Figure 1d and Extended Data Figure 1a). According to Ohm's law and Kirchhoff's current law, if the $j$th bottom bar is grounded, it has the current output of $I_j = \sum_i^m V_i G_{i,j}$ ($j = 1, \ldots n$), and the $(n+1)$th bottom bar has the output current of $I_{ref} = \sum_i^m V_i G_{ref}$. The weighted summation has been completed in the sense that $y_j^l = \frac{I_j - I_{ref}}{V_0 G_0}$. However, the currents $I_j$ and $I_{ref}$ will not be measured, nor do we need to explicitly calculate the membrane potential $y_j^l$.

**Stochastic binarization of the forwarding signals in hardware circuit**

The stochastic binarization of the forwarding signals, i.e., the Bernoulli sampling process, $P(x_j^{l+1} = 1) = z_j^l = \frac{1}{1+\exp(-y_j^l)}$ could be directly implemented in a dedicatedly designed hardware circuit (Figure 1d and Extended Data Figure 1a). First, we add a noise current signal $I_{noise}$ into the output current of the bottom bars of the memristor array $I_j$. We then convert the combined current $I_j + I_{noise}$ and the reference current $I_{ref}$ into voltage signals of $V_j = -R_t(I_j + I_{noise})$ and $V_{ref} = -R_t I_{ref}$, respectively, through trans-impedance amplifiers with $R_t$ being the feedback resistance. The trans-impedance amplifiers also pull the bottom bars into a virtually grounded state. The voltage signals of the trans-impedance amplifiers are fed to a comparator that outputs logic 1 voltage level (i.e., $V_{DD}$) when $V_j < V_{ref}$ and logic 0 voltage level (i.e., 0V) otherwise. Extended Data Figure 2a and 2b show the typical behaviors of such a circuit, where the noise current is obtained by amplifying the thermal noise of resistors. We sample the comparator's output voltage $V_{out_j}^l$ by a flip-flop and the sampled value is taken as the input signal $x_j^{l+1}$ for the next layer.

If $I_{noise}$ follows normal distribution $I_{noise} \sim N(\mu, \sigma^2)$, we have:

$$P(x_j^{l+1} = 1) = P(V_j < V_{ref})$$
$$= P(I_j + I_{noise} > I_{ref})$$
$$= P(I_{noise} > I_{ref} - I_j)$$
$$= 1 - \int_{-\infty}^{I_{ref} - I_j} \frac{1}{\sigma\sqrt{2\pi}} \exp\left(-\frac{(x-\mu)^2}{2\sigma^2}\right) dx$$
$$= \frac{1}{2}\left[1 + \text{erf}\left(\frac{(I_j - I_{ref}) + \mu}{\sigma\sqrt{2}}\right)\right]$$
$$= \frac{1}{2}\left[1 + \text{erf}\left(\frac{y_j^l V_0 G_0 + \mu}{\sigma\sqrt{2}}\right)\right]. \tag{3}$$

Note that Eq. 3 is also a sigmoid function. For the given read voltage $V_0$ and scaling factor $G_0$, Eq. 3 can closely approximate the desired logistic function of $z_j^l = \frac{1}{1+\exp(-y_j^l)}$ if the expectation and the standard deviation of the current noise, i.e., $\mu$ and $\sigma$, could be appropriately chosen.

For typical values $V_0 = 0.1$ V and $G_0 = 10^{-6}$ S, we simulate the circuit behaviors with different noise current levels. The simulation shows that when $\mu = 0$ μA and $\sigma = 0.175$ μA, the circuit behavior captures the logistic function well (Extended Data Figure 2c). Similar stochastically activated neuronal behaviors have been recently reported exploiting various types of noise sources[27,40,41]. When $\mu = 0$ μA, changing of noise levels is equivalently scaling the membrane potential $y_j^l$ in the logistic function with a prefactor $a$, that is $\frac{1}{1+\exp(-a*y_j^l)}$. Since the prefactor won't affect the overall learning performance (Extended Data Figure 5a-5e), a coarse control of the noise level would be sufficient.

Also, note that Eq. 3 is the cumulative density function of the normally distributed current noise $I_{noise}$, with respect to the current difference $I_j - I_{ref}$. We have confirmed that the proposed neural network algorithms work well as long as the activation function is of the sigmoid type (Extended Data Figure 5). Thus, we are not constrained to approximate the logistic function. In other words, we are not limited to using the noise source that follows a normal distribution, any type of noise can be used for the stochastic binarization, due to the simple fact that the cumulative density function is always of the sigmoid type.

**Stochastic binarization of the derivatives in hardware circuit**

After the logistic function $z_j^l = \frac{1}{1+\exp(-y_j^l)}$ has been stochastically binarized, the stochastic binarization of its derivative $P\left(\frac{\partial z_j^l}{\partial y_j^l} = 1\right) = z_j^l(1 - z_j^l)$ can be easily implemented. As shown in Figure 1e and Extended Data Figure 1b, we use two flip-flops to conduct two independent Bernoulli sampling processes on the comparator's output $V_{out_j}^l$ and the sampled values (logic signal $A$ and $B$) are processed by a logic gate. The logic gate is composed of a NOT gate and an AND gate. The NOT gate reverses the second sampled value ($\bar{B}$), while the AND gate outputs the logical conjunction of the first sampled value $A$ and the reversed second sampled value $\bar{B}$, that is, $A \cap \bar{B}$. The logic gate outputs 1 only when the first sampled value is 1 and the second sampled value is 0. Thus, assuming that $I_{noise}$ follows normal distribution $I_{noise} \sim N(\mu, \sigma^2)$, we have:

$$P\left(\frac{\partial z_j^l}{\partial y_j^l} = 1\right) = P(A = 1)[1 - P(B = 1)]$$

$$\begin{aligned}
&= P(V_j < V_{ref})[1 - P(V_j < V_{ref})] \\
&= P(I_{noise} > I_{ref} - I_j)P(I_{noise} < I_{ref} - I_j) \\
&= \tfrac{1}{2}\left[1 + \text{erf}\left(\tfrac{y_j^l V_0 G_0 + \mu}{\sigma\sqrt{2}}\right)\right] * \tfrac{1}{2}\left[1 - \text{erf}\left(\tfrac{y_j^l V_0 G_0 + \mu}{\sigma\sqrt{2}}\right)\right] \\
&\approx \tfrac{1}{1+\exp(-y_j^l)}\left[1 - \tfrac{1}{1+\exp(-y_j^l)}\right] \\
&= z_j^l(1 - z_j^l). \quad\quad\quad\quad\quad\quad\quad\quad\quad\quad\quad\quad\quad\quad\quad\quad (4)
\end{aligned}$$

Extended Data Figure 2d shows the comparison between the desired derivative probability and circuit behaviors of several different noise levels. When $\mu = 0$ µA and $\sigma = 0.175$ µA, the memristor array hardware's behavior can well resemble the desired activation's derivative.

The derivation of Eq. 4 relies on two assumptions: i) the similarity between the logistic function and the cumulative density function of the normal distribution; ii) the simple expression of the derivative of the logistic function $z_j^l(1 - z_j^l)$ where the argument of the logistic function, $y_j^l$, is not explicitly needed. However, these two assumptions should not be the priorities for us to use the designed hardware circuit for the stochastic binarization of the activation function's derivative. As we have confirmed in Extended Data Figure 5f-5k, the binary stochastic learning algorithm would work nicely as long as the activation is of sigmoid type (has an "S" shape) and the "function derivative" is bell-shaped. The "function derivative" needs not be the exact derivative of the activation function. For any type of noise source, if the membrane potential $y_j^l \propto I_j - I_{ref}$ is small enough, the $P(A = 1) = P(I_{noise} > I_{ref} - I_j)$ approximates 0, and so does $P\left(\frac{\partial z_j^l}{\partial y_j^l} = 1\right)$. While, if the membrane potential is large enough, the $1 - P(B = 1) = P(I_{noise} < I_{ref} - I_j)$ approximates 0, and so does $P\left(\frac{\partial z_j^l}{\partial y_j^l} = 1\right)$. When the membrane potential is near the expectation of the noise sources, that is, $I_{noise} \approx I_{ref} - I_j$, both $P(A = 1)$ and $1 - P(B = 1)$ approximate 0.5, and $P\left(\frac{\partial z_j^l}{\partial y_j^l} = 1\right)$ takes the maximal value of 0.25. Thus, a "bell" shape of the "function derivative" is well produced.

Eq. 4 has the maximal value of 0.25. It can be linearly scaled to have the maximum values of 0.5 and 1 (Extended Data Figure 5b and 5c) by repeating the circuit behavior of Eq. 4 for two times and four times, respectively, and taking the logical disjunction of the outputs of these repetitions as the stochastic binarization of the derivative. The probability of the logical disjunction is the union of the individual implementation of Eq. 4.

Note that, the two logic levels $A$ and $B$ should be sampled independently, which is now realized by sampling the comparator's output $V_{out_j}^l$ in two clock cycles. Otherwise, the joint probability in Eq. 4 is not valid. It should also be reminded that these two sampling processes should both be independent of the sampling of the forwarding signals in Eq. 3. Otherwise, the learning rule denoted by Eq. 2 would not succeed since the equivalence to the Eq. 1 by the law of total expectation is invalid. Thus, the forwarding pass needs at least three clock cycles to obtain the stochastic binarizations of both the forwarding signals and the derivatives (top right inset in Extended Data Figure 1).

**Multiple methods of realizing the binary stochasticity in hardware**

The stochastic binarization for the forwarding signals and activation derivatives can be achieved in multiple ways, in addition to our proposal (Extended Data Figure 2). For instance, the intrinsic noise in the crossbar array of memristors[27], and the stochastic nature of the switching process of a diffusive memristor[40] or a magnetic tunnel junction[41] can also be exploited to obtain the stochastic binary output in the neuron.

**Limitation of previous utilizations of the hardware stochasticity**

The binary stochasticities that can be provided by the intrinsic noise or stochastic nature make them ready to be employed in Hopfield-type neural networks, such as finding the global minima in constraint satisfaction problems[27,41–43] or learning through contrastive divergence in restricted Boltzmann machines[44,45]. However, for deep learning with the error backpropagation and gradient descent rule, the jigsaw puzzle of the in-situ learning independent of the von-Neumann architecture within a neuromorphic system has not been completed. For instance, a neural sampling machine with stochastic synapses of ferroelectric field effect transistors was recently reported for the learning and inference of a fully-connected neural network[46]. While the forwarding signals are stochastically binarized, the error backpropagation and the gradient of the weights are calculated in a traditional high-precision scheme[47]. Thus, only the inference and the forward pass in the learning stage are accelerated. The error backpropagation still needs complex peripheral circuits, and the weight update operation needs to be precisely controlled. We complete the jigsaw puzzle by introducing the stochastic binarization of the activation derivatives and the sign of backpropagating errors. They helped to accelerate the error backpropagation and weight update in the hardware implementation of deep learning.

**Backward weighted summation by the memristor crossbar**

The same memristor crossbar is used for the backward weighted summation of the backpropagating errors $\delta x_i^l = \sum_j \delta y_j^l w_{i,j}^l$. The ternary valued errors of member potentials $\delta y_j^l$ (taking values of "-1", "0", and "1") are represented by signals of $-V_{DD}$, high impedance, and $V_{DD}$, respectively. Level shifters are used to convert the voltage signals $-V_{DD}$ and $V_{DD}$ to read voltages with the amplitude of $-V_0$ and $V_0$, respectively, and to convert the high-impedance signal to 0V. Thus, the voltages $V_j^b = V_0 \delta y_j^l$ ($j = 1, \ldots n$) are applied to the $j$th bottom bar of the memristor array (Figure 1f and Extended Data Figure 1c). According to Ohm's law and Kirchhoff's current law, if the $i$th top bar is grounded, it has the current output of $I_i^b = \sum_j^n V_j^b G_{i,j}$ ($i = 1, \ldots m$), and the $(m+1)$th top bar has the output current of $I_{ref} = \sum_j^n V_j^b G_{ref}$. The backward weighted summation has been completed in the sense that $\delta x_i^l = \frac{I_i^b - I_{ref}}{V_0 G_0}$. Similar to the case in the forward-weighted summation, the currents $I_i^b$ and $I_{ref}$ will not be measured, nor do we need to explicitly calculate the errors $\delta x_i^l$.

**Sign operation of the backpropagating errors in the hardware circuit**

The sign operation of the backpropagating error on a neuron $\delta z_j^l = sign(\delta x_j^{l+1})$ is conducted by a current comparator who compares the currents $I_i^b$ with $I_{ref}$ (Figure 1f and Extended Data Figure 1c). The current comparator outputs the positive voltage $V_{DD}$ to represent the signed error 1 when $I_i^b \geq I_{ref}$ and outputs the negative voltage $-V_{DD}$ to represent the signed error -1 otherwise (Figure 1f). In Extended Data Figure 1c, the current operator is implemented by first converting the currents to voltages through trans-impedance amplifiers and then comparing the voltages through a voltage comparator. The trans-impedance amplifiers also pull the top bars into a virtually grounded state.

The multiplication between the error on a neuron and the activation function's derivative $\delta y_j^l = \delta z_j^l \frac{dz_j^l}{dy_j^l}$ can be implemented by a single transistor (Figure 1f and Extended Data Figure 1c), since the error on a neuron $\delta z_j^l$ is binary valued ($-V_{DD}$ and $V_{DD}$ for -1 and 1 respectively), and the derivative $\frac{dz_j^l}{dy_j^l}$ is binary valued (0 and $V_{DD}$ for 0 and 1 respectively). The former is applied to the source/drain of a transistor, while the latter is applied to the gate of the transistor. The product is then presented in the drain/source terminal of the transistor, being $-V_{DD}$, high impedance, and $V_{DD}$ for -1, 0, and 1, respectively.

**The input to the first layer**

The first layer of a deep neural network, or the input layer, receives information directly from the training samples during the learning. To make the first layer compatible with the binary stochastic learning algorithm and the memristive hardware implementation scheme, the input information from the training samples should also be stochastically binarized. The input information is first normalized to the range of 0 to 1 and then binarized to either 0 or 1 through the Bernoulli process.

For instance, in the task of learning the handwritten digits from the MNIST dataset, the raw data are represented in 8-bit unsigned integers: the grayscale of each pixel of the digit image is represented by an integer from 0 to 255 with 0 being fully black and 255 fully white (Figure 2a). These integers are scaled to values between 0.0 to 1.0 by dividing the integer with the largest number 255. The scaled pixel values are used as the probability of the corresponding input nodes taking value 1 instead of 0. In the control experiment of high-precision learning, the scaled pixel values are directly used as the values of input nodes.

**The activation of the output layer**

The output layer of the neural networks should also be specially cared for. In the demonstrations of the classification tasks (Figure 2 and Figure 4), the SoftMax activation function and cross-entropy loss are used for the output layer. The activation of the output layer is given by $z_j = \frac{\exp(y_j)}{\sum_k \exp(y_k)}$, where $y_j$ is the membrane potential of the neurons in the output layer and $j = 1, \ldots, 10$ for the classification tasks in this work. The cross-entropy loss given by $L = -\sum_j t_j \log(z_j)$, where $t_j$ ($j = 1, \ldots, 10$) is the target membrane potential of the neuron given by the corresponding label of the training sample and encoded in a one-hot format. The strategy of deep learning is to reduce cross-entropy loss by adjusting all the weights in the neural network.

Traditionally, the errors of the membrane potential of the neurons in the output layer ($\delta y_j$), that is, the blames of the resulting cross-entropy loss that could be assigned to the membrane potentials, is given by the gradient of the cross-entropy loss with respect to the membrane potential of each neuron, reading,

$$\delta y_j = \frac{\partial L}{\partial y_j} = \sum_k \frac{\partial L}{\partial z_k} \frac{\partial z_k}{\partial y_j} = z_j - t_j. \quad (5)$$

The errors of the neurons $\delta y_j$ are then backpropagated in the deep neural network. Note that the value of the cross-entropy loss $L$ is not needed for calculating $\delta y_j$.

In binary stochastic learning, the activation of the output layer $z_j$ acts as the probability of the output neuron to be activated, that is $P(z_j^B = 1) = z_j$, where $z_j^B$ is the binary state of the output neuron. The error of the membrane potential $\delta y_j$ for backpropagating is replaced by

$$\delta y_j = z_j^B - t_j. \quad (6)$$

Since the target outputs $t_j$ are also binary valued in one-hot format, the errors $\delta y_j$ are ternary valued (-1, 0, or 1), the same as the errors in previous layers.

The calculation of the SoftMax activation and the Bernoulli sampling process are currently implemented in the software. However, they can also be implemented by hardware circuits with the help of noise sources without explicitly obtaining the exact value of the membrane potential $y_j$ and calculating the SoftMax activation function. To achieve this, the winner-take-all mechanism with inhibitory connections among neurons should be employed, similar to the case in biological systems[48].

**Fully-connected neural network**

The neural network shown in Figure 2a consists of three fully-connected layers with the size of 784 × 500, 500 × 200, and 200 × 10. The 784 input nodes correspond to 784 (28 × 28) pixels of one MNIST training sample, and the 10 output nodes correspond to the 10 classes of digits.

In Extended Data Figure 5a-e, we test the effect of the shape parameter $a$ of the logistic activation function $z_j^l = \frac{1}{1+\exp(-ay_j^l)}$ in various signal/error/derivative handling methods. The forwarding signals (F) can be in high-precision (HP) or stochastic binary (S) format; the backpropagation errors (E) can be in high-precision (HP) or signed (S) format; the derivative (D) can be in high-precision (HP) or stochastic binary (S) format, as shown in Extended Data Figure 5e. In stochastic binarization, the "probability" that is higher than 1 is truncated (for instance, in Extended Data Figure 5d). The cases labeled by "F:HP, E:HP, D:HP" correspond to the high-precision learning in the main text, whereas the cases labeled by "F:S, E:S, D:S" correspond to the binary stochastic learning. The highest learning performance (>99% recognition accuracy) is achieved when the forwarding signals and the derivatives are binarized and the backpropagating errors are in high precision, i.e. "F:S, E:HP, D:S" with the shape parameter $a$ being 8. However, this case is not fully compatible with the designed memristive hardware circuit.

We also test the effect of other types of activation functions in Extended Data Figure 5f-k. Note that, for the rectifying linear unit activation function (Extended Data Figure 5f), the learning algorithms with high-precision forwarding signals work well, whereas the learning algorithms with binary stochastic forwarding signals are disruptive. For Extended Data Figure 5i and 5j, the "derivative functions" that are used in the learning are faked (they are not the derivative function of the activation functions). In these two cases, the learning algorithms still work well.

We use the logistic activation function with the shape parameter $a$ being 4 (Extended Data Figure 5c) for the learning results shown in Figure 2a, Figure 3, Extended Data Figure 3, and Extended Data Figure 4.

All the learnings use a batch size of 100 and a fixed learning rate $\eta = 0.1$. By default, we train the neural network for 1000 epochs. No other learning performance enhancement techniques, such as dropout, batch normalization, and data preprocessing, are used.

**Deep convolutional neural network**

The learning algorithms for the fully-connected layers could be directly transferred to the convolutional layers with minor changes. The weighted summation in the fully-connected layer between the one-dimensional input vector and the two-dimensional synaptic matrix becomes the convolution between two/three-dimensional feature maps and convolutional kernels. For the convolutional layer without being followed by a pooling layer, the transmission of the forwarding signals, the derivatives, and the backpropagating errors are the same as in the fully-connected layers. For the convolutional layer followed by a pooling layer, it should be decided where the activation function and the Bernoulli sampling should be performed. In this work, we used the max-pooling layer, with the following strategy: i) the activation function is performed on the membrane potentials of the neurons in the convolutional layers; ii) the max-pooling is performed on the activation; and iii) the Bernoulli samplings of both the forwarding signals and the derivatives are performed on the neurons in the pooling layers.

In Figure 4a for learning and recognizing the handwritten digits in the MNIST dataset, the deep convolutional neural network consists of two convolutional layers, two max-pooling layers, and one fully-connected layer. The first convolutional layer has eight filters that are using kernel sizes of $9 \times 9$. The second convolutional layer has 12 filters that are using kernel sizes of $5 \times 5$. Both convolutional layers are followed by non-overlapping max-pooling layers with pooling sizes of $2 \times 2$. The fully-connected layer uses a $108 \times 10$ synaptic weight matrix. We use the logistic activation function with the shape parameter $a$ being 4 for regular layers, except the output layer which uses the SoftMax activation function. The neural network is trained using a batch size of 100 and a fixed learning rate of 0.1.

In Figure 4f for learning and recognizing images in the CIFAR-10 dataset, the input is of dimensions of $32 \times 32 \times 3$, with 3 being the RGB channels of the colored images. The deep convolutional neural network uses a visual geometry group (VGG) style, consisting of six convolutional layers, three max-pooling layers, and three fully-connected layers. All convolutional layers have the same kernel size of $3 \times 3$ with padding on the edges. All max-pooling layers use the non-overlapping pooling windows with the same size of $2 \times 2$. There is one pooling layer following two convolutional layers. The sizes of the feature maps and the channels for each layer are given in Figure 4f. The neural network is trained using a batch size of 100 and a fixed learning rate of 0.01.

## Modeling of the synaptic behaviors of the memristors

An empirical model capturing the synaptic behavior of long-term potentiation and long-term depression under identical pulses is used to simulate the synaptic plasticity of the analog memristors with non-idealities[45]. This model considers the on/off ratio, the non-linearities ($\alpha_p$ and $\alpha_d$), the asymmetry between potentiation and depression, and the write variations. The median conductance changes (without cycle-to-cycle write variations) for a memristor device with conductance $G_{ij}$ under potentiation pulses and depression pulses can be written as,

$$\overline{\Delta G_{pot}} = [\frac{G_{max}-G_{min}}{1-e^{-\alpha_p}} - (G_{ij} - G_{min})](1 - e^{-\alpha_p/N_p}), \qquad (7)$$

and,

$$\overline{\Delta G_{dep}} = -[\frac{G_{max}-G_{min}}{1-e^{-\alpha_d}} - (G_{max} - G_{ij})](1 - e^{-\alpha_d/N_d}), \qquad (8)$$

respectively. Here, $N_p$ and $N_d$ are the numbers of pulses needed to fully potentiate and fully depress the memristor devices, respectively, and $\alpha_p$ and $\alpha_d$ are the non-linearities of weight updates in the potentiation and depression phases, respectively.

The cycle-to-cycle write variations are modeled by adding a Gaussian distribution to the conductance change with its standard deviation proportional to the median conductance change from Eq. 7 or Eq. 8,

$$\Delta G \sim \mathcal{N}[\overline{\Delta G}, (\gamma \overline{\Delta G})^2], \qquad (9)$$

where $\overline{\Delta G} = \overline{\Delta G_{pot}}$ for potentiation pulses, $\overline{\Delta G} = \overline{\Delta G_{dep}}$ for depression pulses, and $\gamma$ is a parameter controlling the cycle-to-cycle variations. For the synaptic behavior in Figure 5b, the values of the parameters are: $G_{max} = 25$ uS, $G_{min} = 0.1$ uS, $N_p = N_d = 100$, $\alpha_p = 1$, $\alpha_d = 2$, and $\gamma = 2$. Note that, due to the write variation, the sign of the actual conductance change $\Delta G$ has a large chance to be in the opposite direction of the desired change.

## Estimation of the energy consumption and on-chip footprint

The energy consumption is estimated in terms of the elementary MAC operations in neural networks (Extended Data Table II).

Implemented in CMOS technology, i.e., using the central processing units (CPUs), graphical processing units (GPUs), or other dedicated application-specific integrated (ASIC) chips, the MAC operations for traditional high-precision learning are conducted typically using 32-bit floating point (FP32) numbers. Each MAC operation consists of the multiplication between two FP32 numbers and the addition of the production to another FP32 number. In a 45 nm CMOS technology node and at 0.9 V supply voltage, this MAC operation[39] consumes a power of 3.7 pJ + 0.9 pJ = 4.6 pJ. In the stochastic binary learning algorithm, the multiplication becomes the production of a Boolean number (0 or 1) and an FP32 number, which needs not to be conducted explicitly. If the Boolean number is 1, the MAC operation only requires the addition of the FP32 number on another FP32 number. If the Boolean number is 0, the addition is not needed. Conservatively, we assume each MAC operation needs one addition between two FP32 numbers, thus the consumed power being reduced to 0.9 pJ. When the weight is quantized to integers, the MAC operation degrades to addition between two integers. For INT8 weights, each MAC operation consumes at most 0.03 pJ. It is convenient to assume that the energy consumption of the addition of two integers is proportional to the bit-width of the integers. Thus, the MAC operations in INT4 weights and ternary valued weights (~1.5 bits) are estimated to consume a power of 15 fJ and 5.6 fJ, respectively.

The crossbar array of the memristors performs all MAC operations in one vector-matrix multiplication parallelly. For instance, the crossbar array with a typical size of 128-by-128 performs $128 \times 128 = 16,384$ MAC operations, parallelly. A macro core of such an array[10] that processes 1-bit input and senses the output currents of the array with analog-to-digital convertors consumes a power of 371.89 pJ. To implement the high-precision learning algorithm, sufficient input accuracy (e.g., 8-bit) is needed for acceptable degradation of the learning performance. There are also shift & adder circuits in the macro core to shift and accumulate the bit-wised MAC results. Thus, each effective MAC operation consumes a power of $\frac{371.89 \text{ pJ} \times 8}{16,384} = 0.18$ pJ. Using our stochastic binary learning algorithm, the analog-to-digital converters and shift & adder are not needed, thus parallel MAC operations consume only 29.23 pJ. Additionally, only 1-bit input is needed. Thus, each MAC operation consumes a power of $\frac{29.23 \text{ pJ}}{16,384} = 1.8$ fJ. The expense induced by trans-impedance amplifiers, the comparators, and the flip-flops in the proposed hardware implementation (Extended Data Figure 1) is not counted in the comparison. They are performing the calculation of the activation function which is done in the digital domain in the benchmark work[10]. The energy efficiency for the MAC operation is approximated to be 556 TOP$s^{-1}W^{-1}$.

The footprint of the implementation circuit would also be greatly reduced. To implement the high-precision learning algorithm, the parallel $128 \times 128 = 16,384$ MAC operations need an on-chip area of 63801.92 um$^2$. Assuming each input bit takes 50 ns, the area efficiency for 8-bit input is calculated as $\frac{16,384}{8*50ns*63801.92um^2} = 641.99$ GOP$s^{-1}mm^{-2}$. Excluding the analog-to-digital converters and shift & adder, the chip area reduces to 8824.3 um$^2$. The effective area efficiency is then $\frac{16,384}{50ns*8824.3um^2} = 37.13$ TOP$s^{-1}mm^{-2}$, reduced by 57.8 times.

**Periodical carry to accumulate the ternary gradient**

The essential of the periodical carry is to accumulate the gradient of the loss function to the weight in a separated memory cell and update the synaptic weight periodically and step-wisely. It is efficient in compensating for the nonlinearity and fluctuation of weight changes[11–14]. Traditionally, the calculation and accumulation of the gradient should be conducted in sufficient precision, for instance, in dedicated capacitors[12] or high-precision digital circuits[13]. In this work, the binarized three factors result in a ternary gradient (valued as -1, 0, or 1), which leads to two-fold benefits. First, the calculation of the gradient for an array of weights, a vector-vector outer product, can be performed parallelly[45]. Second, the gradient is accumulated in a unit step.

**Limitation on neural network depth**

Binary stochastic learning is, however, limited by the depth of the neural network. Learning of a neural network with more than 10 layers is difficult. This is because a long chain of stochastic binarization over initially random weight matrices loses meaningful information. This issue can be partially compensated by pre-training the neural network in high-precision format or increase the batch size, according to our experiences with software simulation. It should be further investigated for extending the proposed algorithms to deeper neural networks for state-of-the-art artificial intelligence applications. However, it should be noted that the human-brain visual ventral pathway mainly consists of several areas, including the retina, lateral geniculate nucleus (LGN), V1, and V4, which correspond to the artificial neural network layers in this work[49]. Within this depth limit, the binary stochastic learning algorithm works well.

# Figures and figure captions

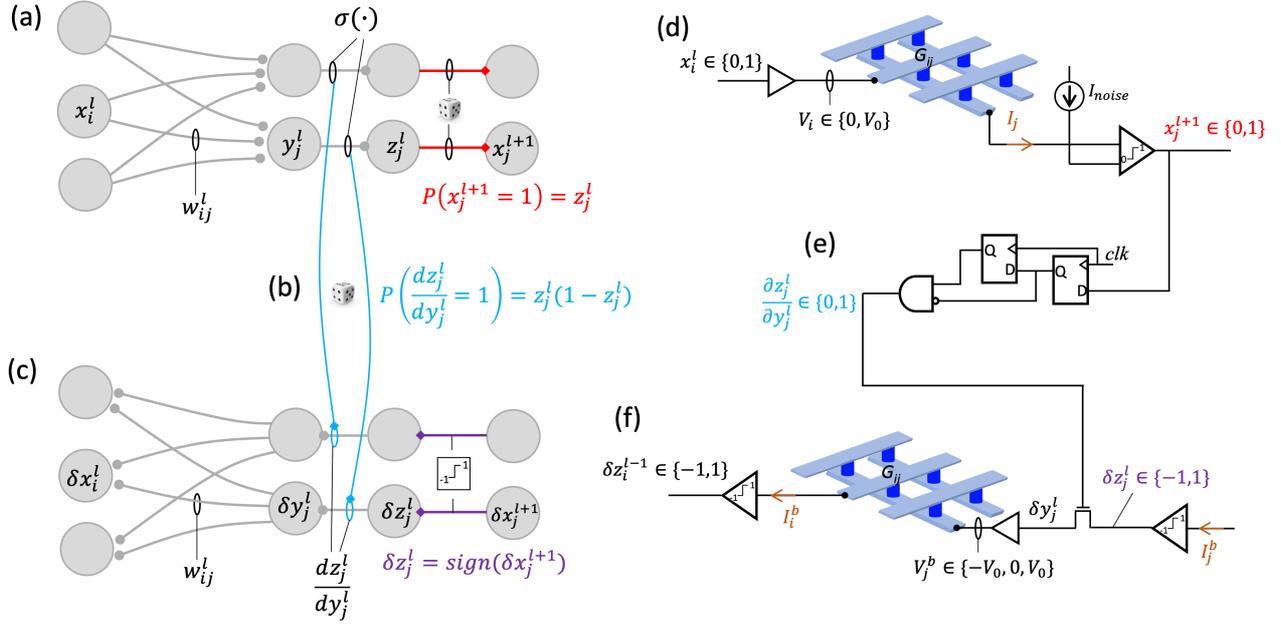

**Figure 1 | Binary stochastic learning algorithm and its hardware implementation. (a)** Stochastic binarization of the signal forwarding. In a typical layer $l$, the neurons are stochastically binarized/activated to be transmitted to the post-layer, i.e., $P(x_j^{l+1} = 1) = z_j^l = \sigma(y_j^l) = \sigma(\sum_i x_i^l w_{ij}^l)$. The input to the weight matrix is binary valued, i.e., $x_i^l \in \{0,1\}$, since the neurons of the pre-layer have also been stochastically activated. **(b)** Stochastic binarization of the activation derivative. The derivative of the activation function is stochastically sampled for the uses in error backpropagation, i.e., $P\left(\frac{dz_j^l}{dy_j^l} = 1\right) = z_j^l(1-z_j^l)$, $\frac{dz_j^l}{dy_j^l} \in \{0,1\}$. **(c)** Error sign backpropagation. Only the signs of the errors from the post-layer are taken to be backpropagated, i.e., $\delta z_j^l = sign(\delta x_j^{l+1})$, $\delta z_j^l \in \{-1,1\}$. The backward input $\delta y_j^l$ to the weight matrix will be ternarily valued since $\delta y_j^l = \delta z_j^l \frac{dz_j^l}{dy_j^l}$. **(d)** The implementation of the stochastically binarized signal forwarding in a crossbar array of memristors. The binary inputs to the array are converted to read voltages by level shifters, and the stochastically binarized outputs are obtained by comparing the output currents of the crossbar array with noise currents, that is, a stochastic activation process. **(e)** The stochastic binarization of the activation derivative by individually sampling the stochastically activated forwarding signal twice and processing the sampled signals through a simple logic gate. **(f)** The implementation of the error sign backpropagation utilizing the same array of memristors. The sign of the error from the post layer is obtained by a comparator and its product with the binarized activation derivative is performed by a transistor. Level shifters are used to apply the ternary valued $\delta y_j^l$ back into the crossbar array.

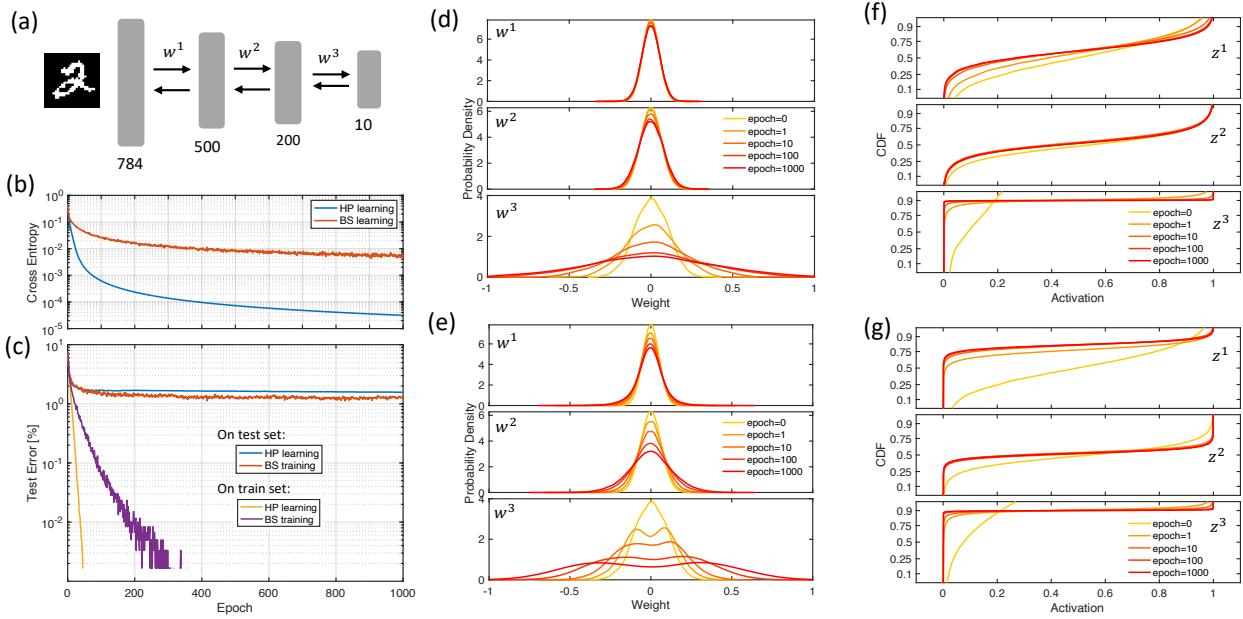

**Figure 2 | High precision learning vs binary stochastic learning.** (**a**) A three-layered fully-connected neural network for testing the learning algorithm on learning the handwritten digits from the MNIST dataset. (**b**) The cross-entropy of the output layer as a function of the learning epoch. (**c**) The test error as a function of training epoch for high precision (HP) learning and the proposed binary stochastic (BS) learning. The binary stochastic learning shows a less overfitting effect on the train set and better recognition performance on unseen handwritten digits from the test set. (**d**) and (**e**), the evolution of the k-s density of weight distributions during high precision learning (d) and binary stochastic learning (e). In binary stochastic learning, more weights in earlier layers have been updated. (**f**) and (**g**), the evolution of the cumulative density function (CDF) of the activations during high precision learning (f) and stochastic learning (g). In binary stochastic learning, the bipartite of activations to the regions near 0 and 1 are more obvious.

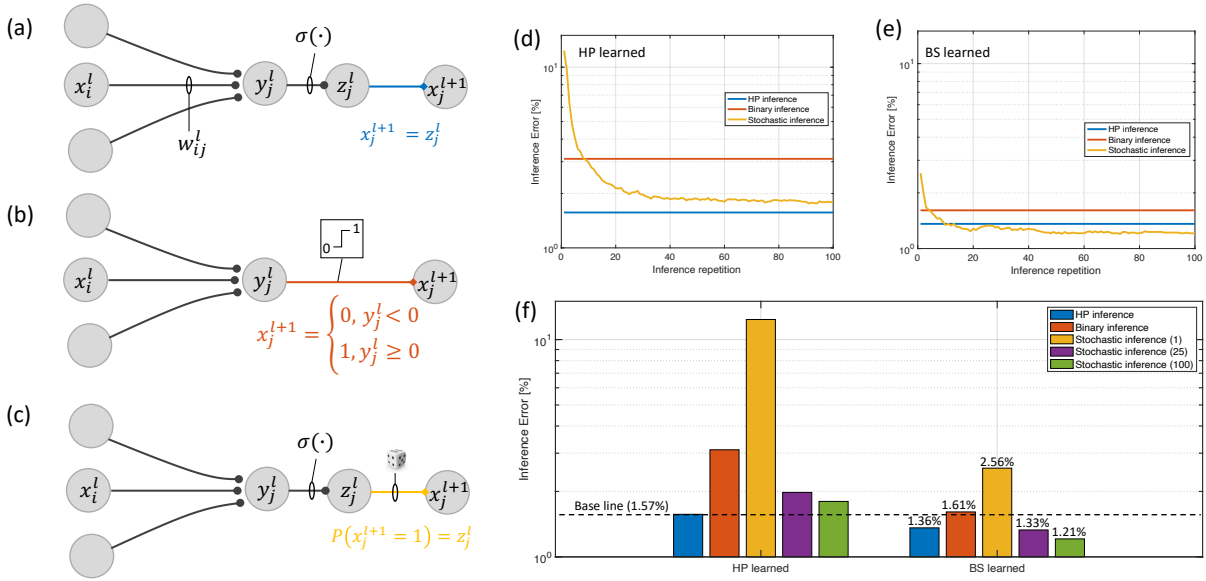

**Figure 3 | Inference methods and inference accuracies. (a)** High precision inference with the input of the post-layer being the activation of the pre-layer. **(b)** Binary inference with deterministic binary activation function. **(c)** Stochastic inference with stochastic binarization within each layer and a majority vote in the output layer. **(d)** Comparison of the inference errors for different inference methods of high precision learned neural network. The high-precision inference is the asymptotic line of the stochastic inference with increasing repetitions. **(e)** Comparison of the inference errors for different inference methods of binary stochastically learned neural network. Better inference performance than high precision inference is obtained for stochastic inference after 10 repetitions. **(f)** Summary of the inference errors comparing high precision learned neural network and binary stochastically learned neural network. The performance (1.57%) in traditional algorithms, i.e., high precision learning and high precision inference, is taken as the baseline. The binary stochastic learning reduces the recognition error by 0.21%, and the stochastic inference reduces the recognition error by 0.15% after 100 repetitions.

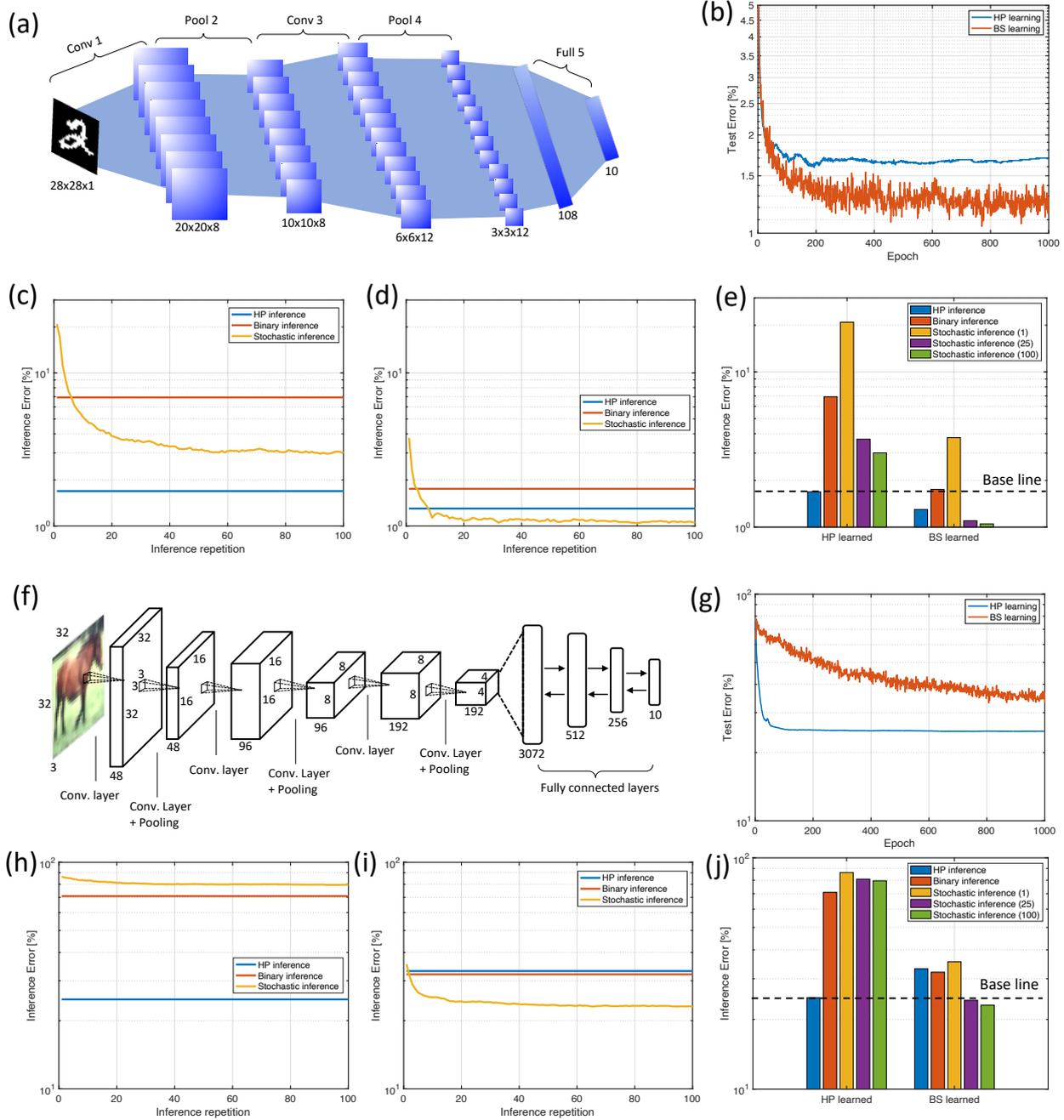

**Figure 4 | The binary stochastic learning and stochastic inference for convolutional neural networks and deeper neural networks. (a)** A five-layer convolutional deep neural network for the MNIST dataset. **(b)** Comparison of the MNIST dataset high-precision inference error rate of each training epoch in two neural networks. Two networks are trained using a high-precision learning algorithm and a binary stochastic learning algorithm, respectively. **(c)** Comparison of the MNIST dataset inference error rate dependency on inference repetition times using different inference methods. The five-layer convolutional neural network is trained using a high-precision learning algorithm. **(d)** Comparison of the MNIST dataset inference error rate dependency on inference repetition using different inference methods. The network is trained using a binary stochastic learning

algorithm. **(e)** Comparison of the test error rates using five different inference methods. The networks are trained using high-precision learning and binary stochastic learning, respectively. **(f)** A convolutional deep neural network for the CIFAR-10 dataset. **(g)** Comparison of the CIFAR-10 dataset high-precision inference error rate of each training epoch in two neural networks. Two networks are trained using a high-precision learning algorithm and a binary stochastic learning algorithm, respectively. **(h)** Comparison of the CIFAR-10 dataset inference error rate dependency on inference repetition times using different inference methods. The five-layer convolutional neural network is trained using a high-precision learning algorithm. **(i)** Comparison of the CIFAR-10 dataset inference error rate dependency on inference repetition times using different inference methods. The neural network is trained using a binary stochastic learning algorithm. **(j)** Comparison of the test error rates using five different inference methods. The networks are trained using high-precision learning and binary stochastic learning, respectively.

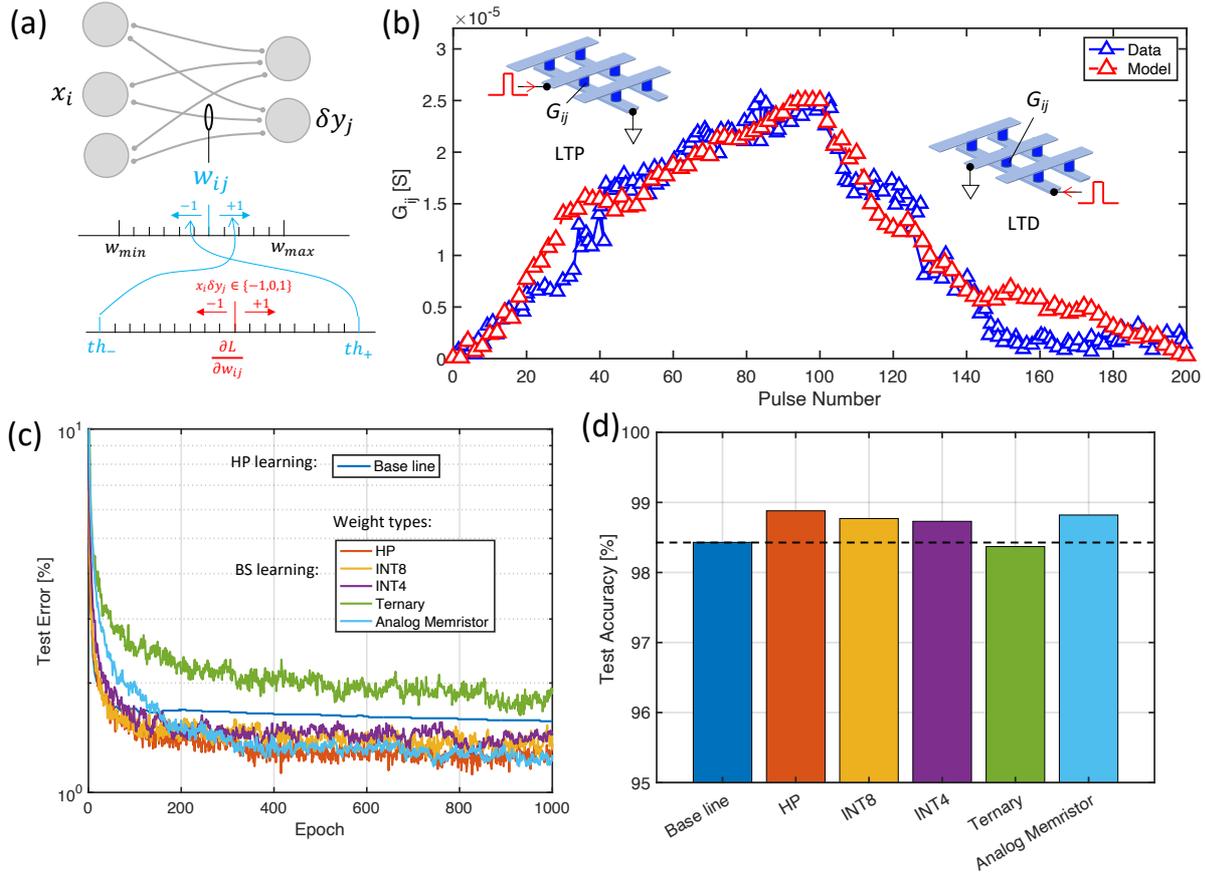

**Figure 5 | Weight quantization and analog weight using memristors. (a)** The integer-styled periodical carry method for training the stochastic neural network in quantized weights represented by integers and analog weights using memristors. **(b)** Typical long-term potentiation (left) and long-term depression behaviors (right) of memristor devices under identical potentiation and depression pulses, respectively. The data are retrieved from the SiGe epitaxial memory[36] and a model is developed to capture the non-linearities and fluctuations of the weight updates. **(c)** The learning curves of the stochastic training using various types of weights compared to the base-line high precision training. **(d)** Summary of the inference accuracy of the stochastically trained using various types of weights compared to the baseline high precision trained neural network. Better than baseline learning performances are obtained for the quantized weights in INT8 and INT4 as well as for the noisy weights in analog memristors.

# Extended Data

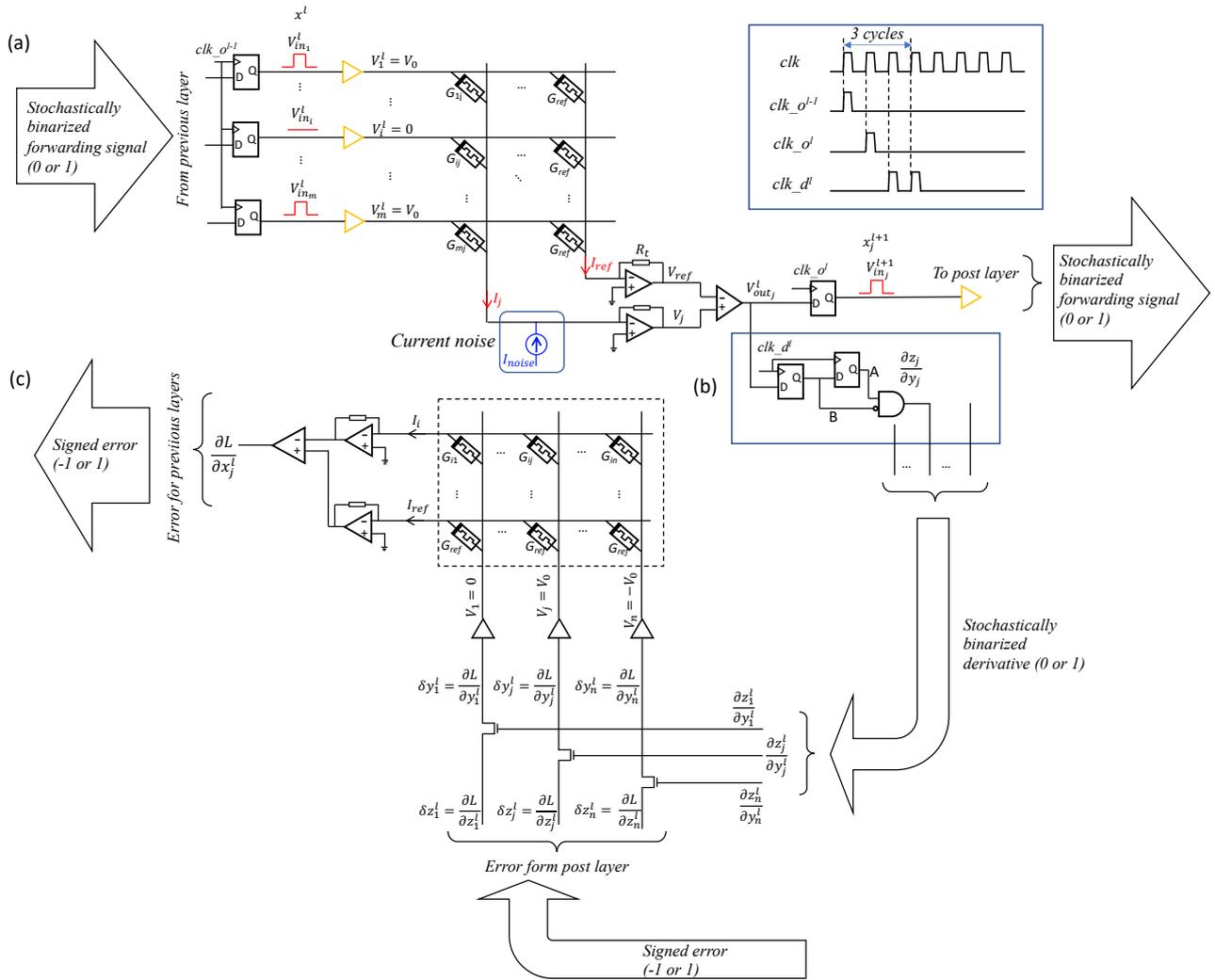

**Extended Data Figure 1 | The hardware implementation of the proposed binary stochastic learning algorithms with a crossbar array of memristors. (a)** The hardware implementation of the binary stochastic sampling for the signals forwarding process. **(b)** The hardware implementation of the binary stochastic of the activation's derivative by two independent Bernoulli sampling processes using flip-flops. **(c)** The hardware implementation of errors backpropagation process of binary (signed) error signals.

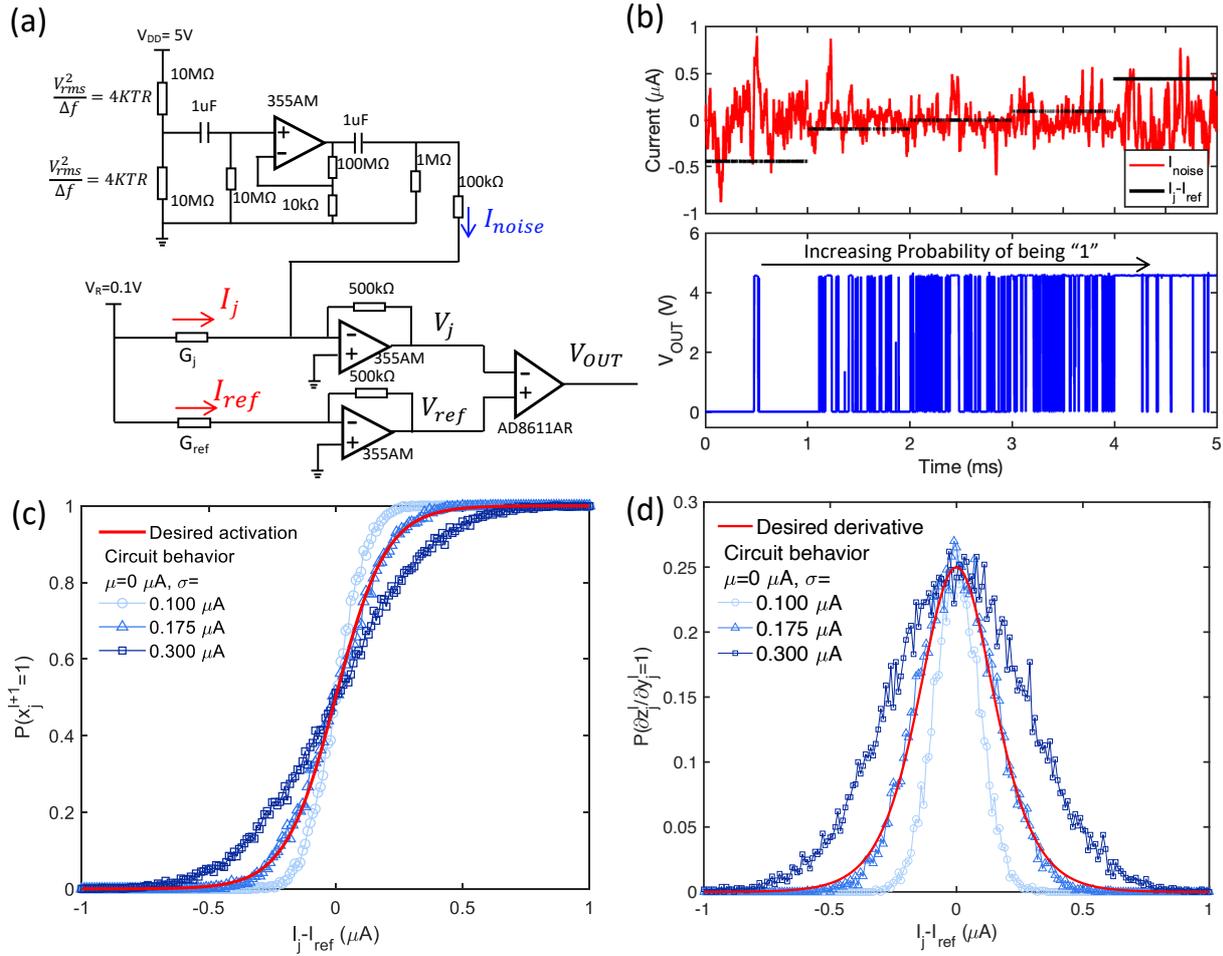

**Extended Data Figure 2 | Comparison of desired stochastic binarization behaviors and memristor array hardware implementation with noise current injection. (a)** A SPICE simulation of the proposed hardware implementation of the binary stochastic neuron. The noise current is provided by amplifying the thermal noise of the resistors. **(b)** The simulated results of the SPICE neuronal circuit showing an increasing probability of the neuron output being "1" with the increase of the current difference between the regular bottom bars ($I_j$) and the reference bottom bar ($I_{ref}$). **(c)** The comparison of the logistic activation function and the circuit behavior with noise. **(d)** The comparison of the activation function derivative and the circuit behavior with noise.

**Extended Data Table I.** Comparison between the binary stochastic (BS) learning and the high precision (HP) learning algorithms.

| Neural network operations / Learning type | HP learning | BS learning |
|---|---|---|
| Forwarding (F) | $x^{l+1} = z^l$ | $P(x^{l+1} = 1) = z^l$ |
| Error backpropagation (E) | $\delta z^l = \delta x^{l+1}$ | $\delta z^l = sign(\delta x^{l+1})$ |
| Derivative (D) of activation | $\dfrac{\partial z^l}{\partial y^l} = z^l(1 - z^l)$ | $P\left(\dfrac{\partial z^l}{\partial y^l} = 1\right) = z^l(1 - z^l)$ |
| Gradient of loss to weight | $\underbrace{\dfrac{\partial L}{\partial w^l}}_{FP} = \underbrace{x^l}_{FP} \underbrace{\dfrac{\partial z^l}{\partial y^l}}_{FP} \underbrace{\delta z^l}_{FP}$ | $\underbrace{\dfrac{\partial L}{\partial w^l}}_{\{-1,0,1\}} = \underbrace{x^l}_{\{0,1\}} \underbrace{\dfrac{\partial z^l}{\partial y^l}}_{\{0,1\}} \underbrace{\delta z^l}_{\{-1,1\}}$ |
| Weight update | $w^l = w^l - \eta \dfrac{\partial L}{\partial w^l}$ | $w^l = w^l - \eta < \dfrac{\partial L}{\partial w^l} >_{batch}$ |

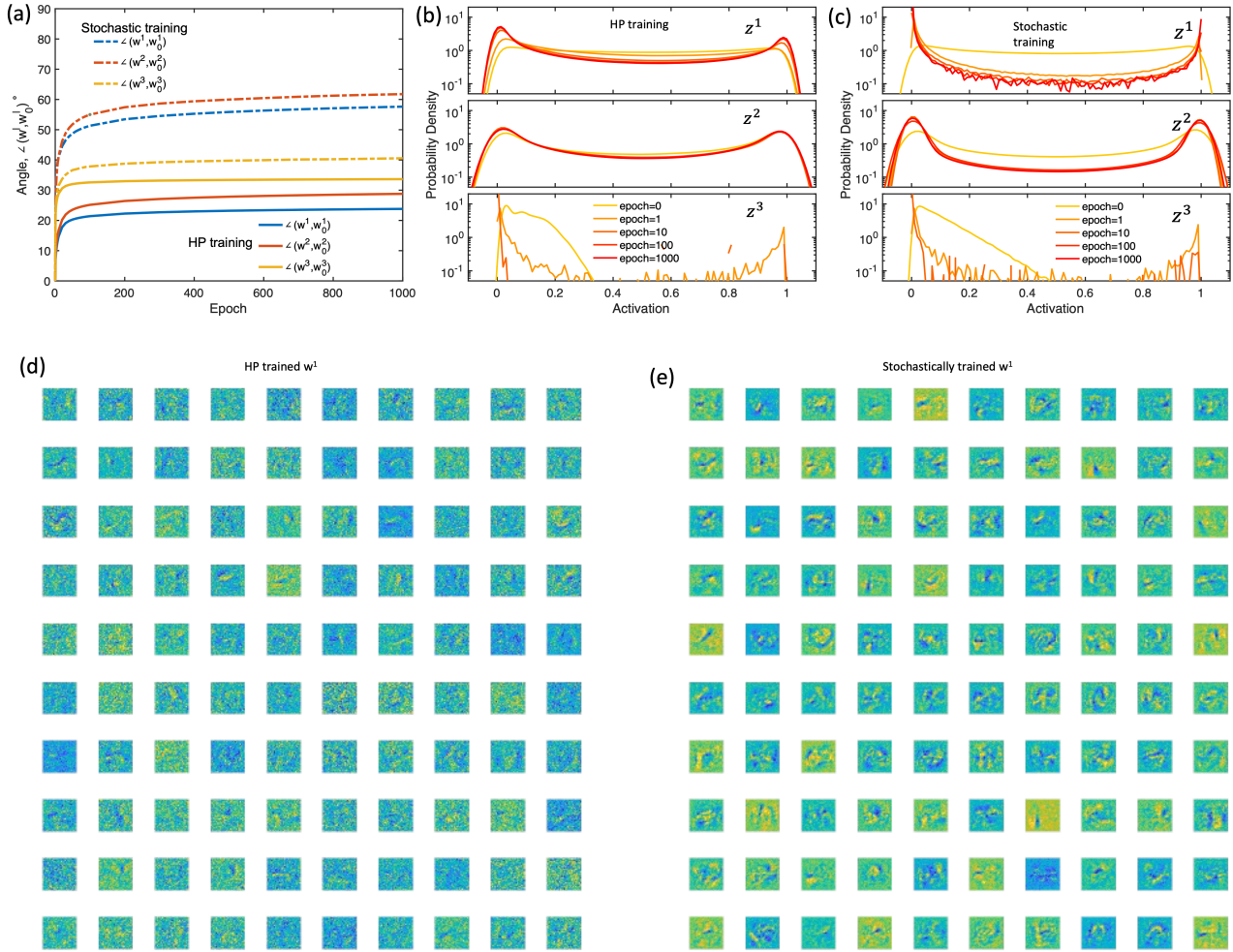

**Extended Data Figure 3 | More details about the evolutions of weight and activation distribution during training. (a)** The evolution of the angle between the learned weights and the initial weights with the progressive of the neural network training. **(b)** The distribution evolution of the neuron activations in the three layers for high-precision learning. **(c)** The distribution evolution of the neuron activations in the three layers for the binary stochastic learning. **(d)** The weight maps of the first layer after the high-precision learning. **(e)** The weight maps of the first layer after the binary stochastic learning. In (d) and (e), each block shows the weights connecting all input neurons (28*28=784) and one output neuron in the first layer. Only the weights connecting to the first 100 output neurons in the first layer are shown here.

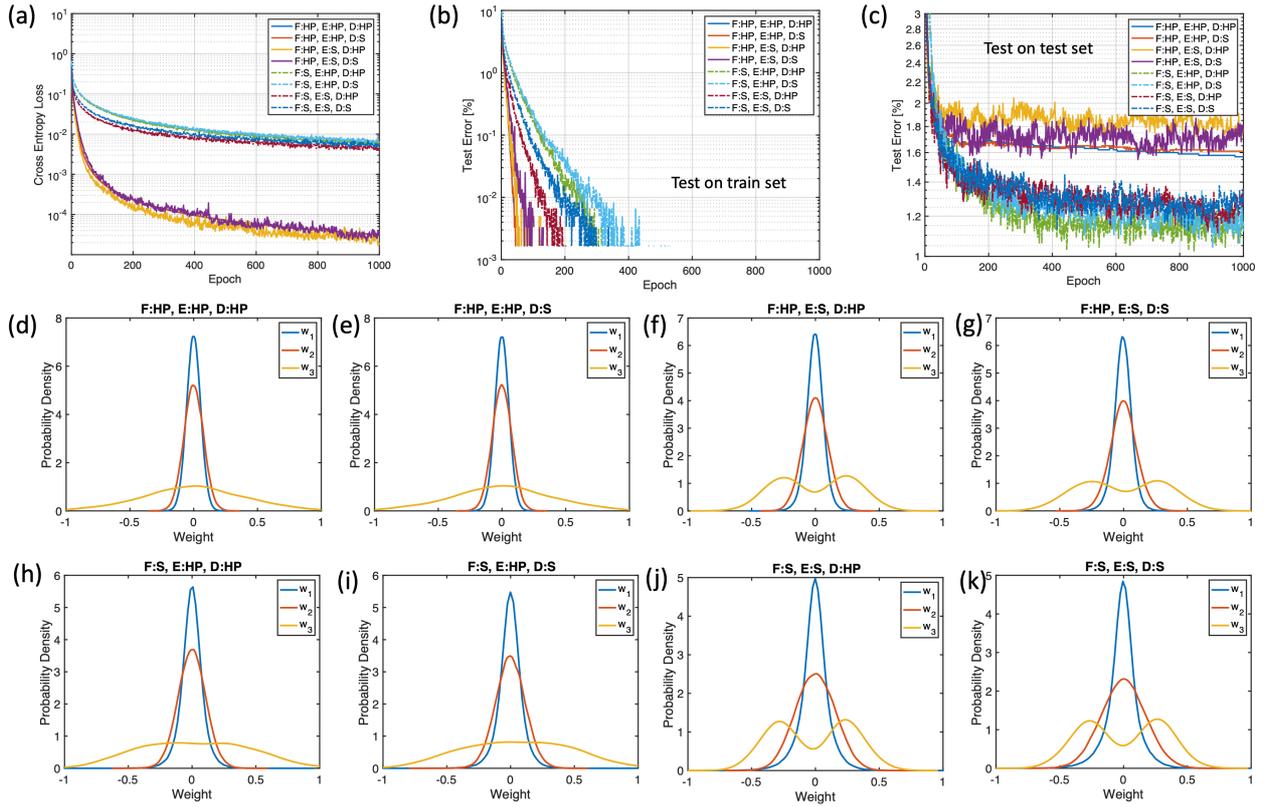

**Extended Data Figure 4 | The comparison of different learning algorithms by combining the proposed elementary neural network operations.** The three elementary operations of signal forwarding (F: being in high precision (HP) or binary sampled (S)), activation derivative (D: being in high precision (HP) or binary sampled (S)) and error backpropagation (E: being in high precision (HP) or signed (S)) can be permutationally combined to generate eight algorithms in total. **(a)** The cross-entropy loss of different learning algorithms. **(b)** The test errors of all eight learning algorithms on the train set. **(c)** The test errors of all eight learning algorithms on the test set. **(d)** The weight distribution of all three layers using a high-precision learning algorithm. The weight distribution of all three layers using the learning algorithm after the learning of **(e)** high-precision signal forwarding, high-precision error backpropagation, and binary stochastic activation derivative, **(f)** high-precision signal forwarding, binary signed error backpropagation, and high-precision activation derivative, **(g)** high-precision signal forwarding, binary signed error backpropagation, and binary stochastic activation derivative, **(h)** binary stochastic signal forwarding, high-precision error backpropagation, and high-precision activation derivative, **(i)** binary stochastic signal forwarding, high-precision error backpropagation, and binary stochastic activation derivative, **(j)** binary stochastic signal forwarding, binary stochastic error backpropagation, and high-precision activation derivative, and **(h)** all three layers using a binary stochastic learning algorithm.

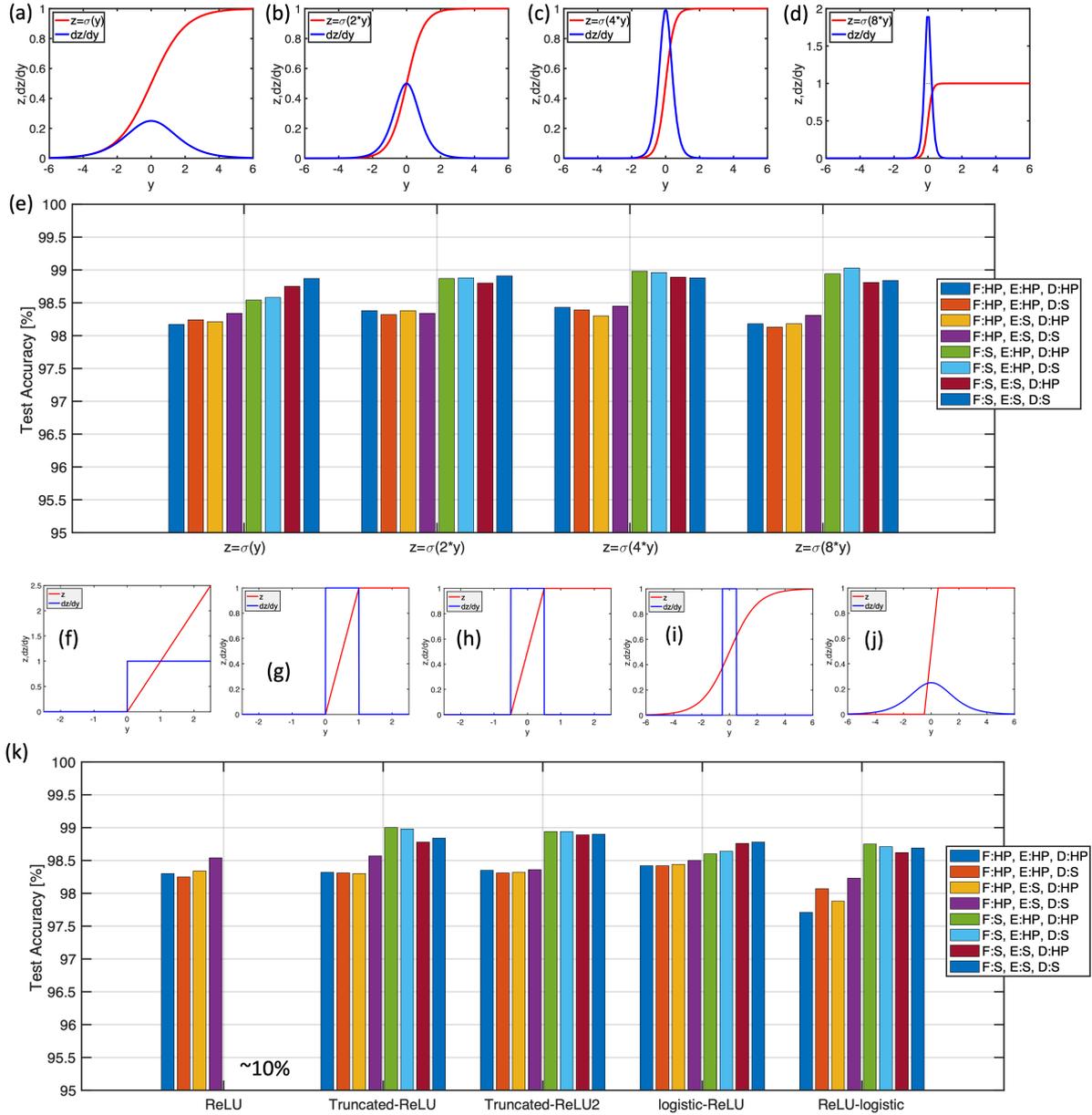

**Extended Data Figure 5 | The effects of different activation functions and activation function derivatives on training performances.** The dependence of the activation function $z = \frac{1}{1+\exp(-a*y)}$ and the activation function derivative $\frac{dz}{dy}$ on $y$: **(a)** $a = 1$; **(b)** $a = 2$; **(c)** $a = 4$; **(d)** $a = 8$. **(e)** The test accuracy of neural networks trained using different activation functions in (a)-(d) permutationally combined with all eight learning algorithms. **(f)** The rectified linear unit (ReLU) function and its derivative. **(g)** A truncated ReLU function (centered at $y = 1/2$) and its derivative. **(h)** A truncated ReLU function (centered at $y = 0$) and its derivative. **(i)** A logistic function and a rectangular function to replace its derivative. **(j)** A truncated ReLU function and a bell-shaped function to replace its derivative. **(k)** The test accuracy of neural networks trained using different activation functions and activation derivatives in (f)-(j) permutationally combined with all eight learning algorithms.

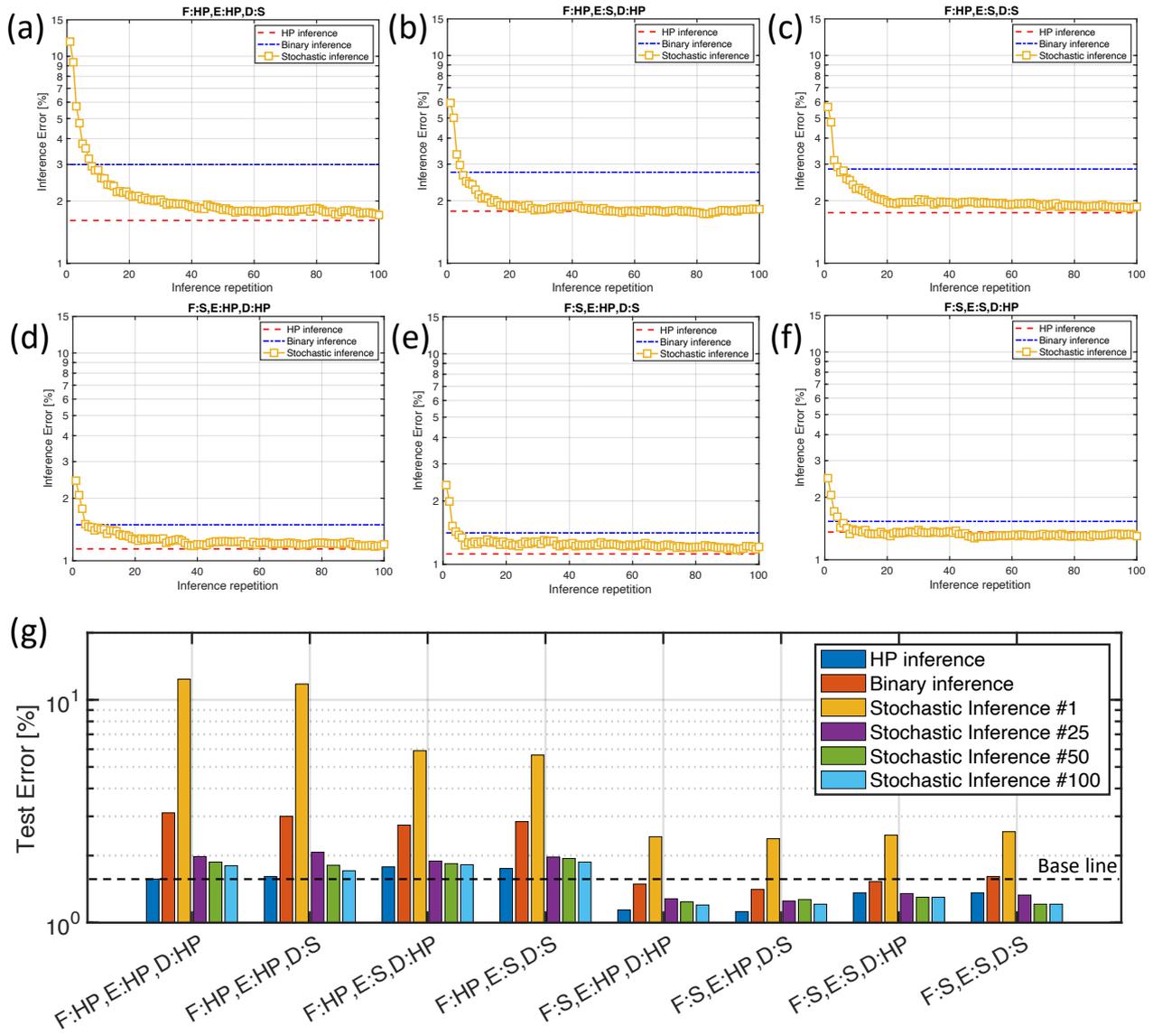

**Extended Data Figure 6 | Inference results for neural networks learned in various learning algorithms. (a)-(f)** Comparison of the inference errors of the three different inference methods for neural networks trained in various learning algorithms. **(g)** Summary of the inference errors for HP inference, binary inference, and stochastic inferences for typical repetition times.

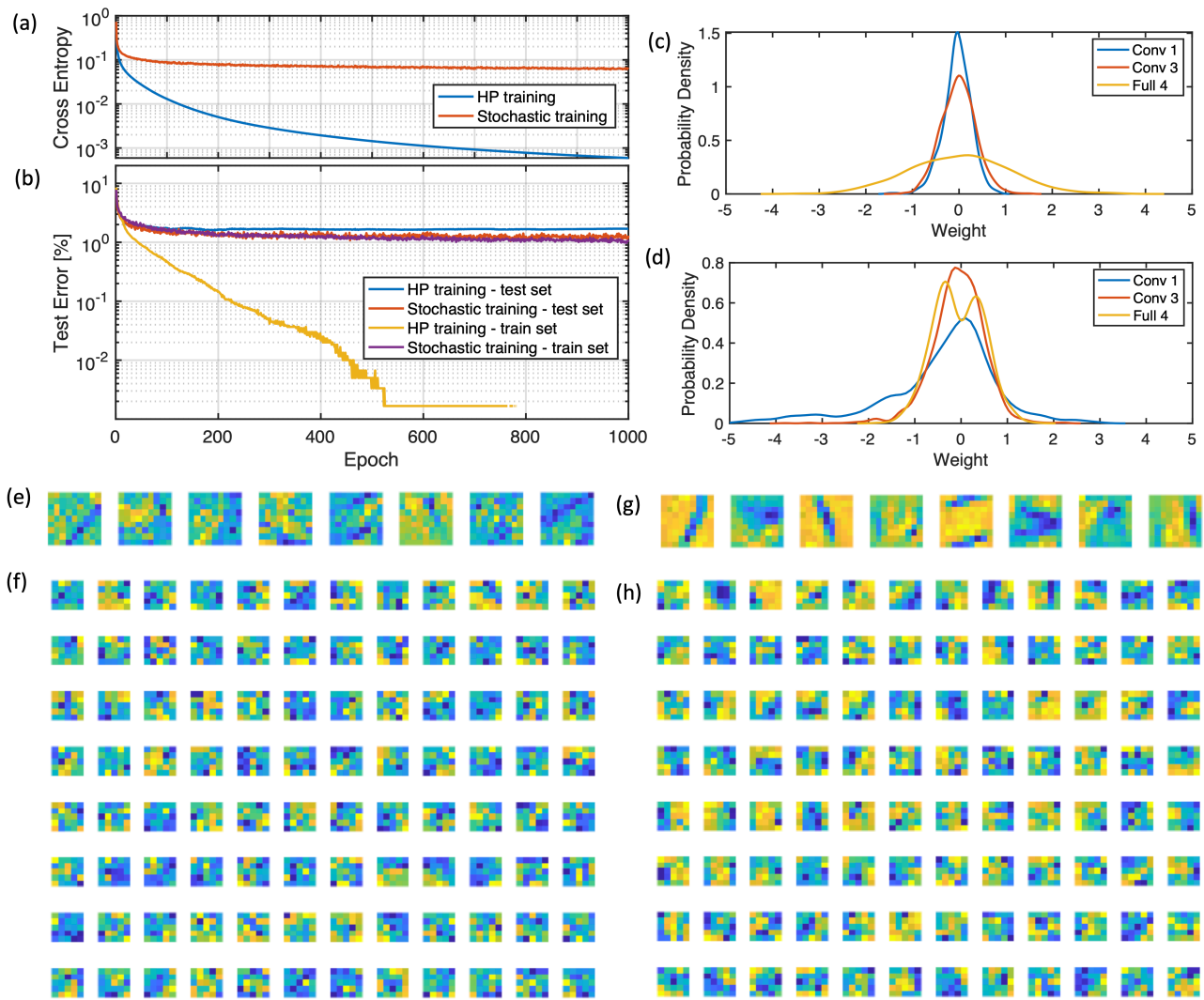

**Extended Data Figure 7 | More data about learning results of convolutional neural network with MNIST dataset. (a)** The cross-entropy loss of the five-layered convolutional neural network (Figure 4a) as a function of the learning epoch. **(b)** The test errors on both the test set and train set during the high precision learning and binary stochastic learning. No overfitting effect is observed in binary stochastic learning. **(c)** The weight distribution of the convolutional layers and fully-connected layers after high-precision learning. **(d)** The weight distribution of the convolutional layers and fully-connected layers after binary stochastic learning. **(e)** and **(f)** The learned convolutional kernels in conv-1 (e) and conv-3 (f) after high-precision learning. **(g)** and **(h)** The learned convolutional kernels in conv-1 (g) and conv-3 (h) after binary stochastic learning. The learned convolutional kernels of stochastically trained neural networks learned more significant features.

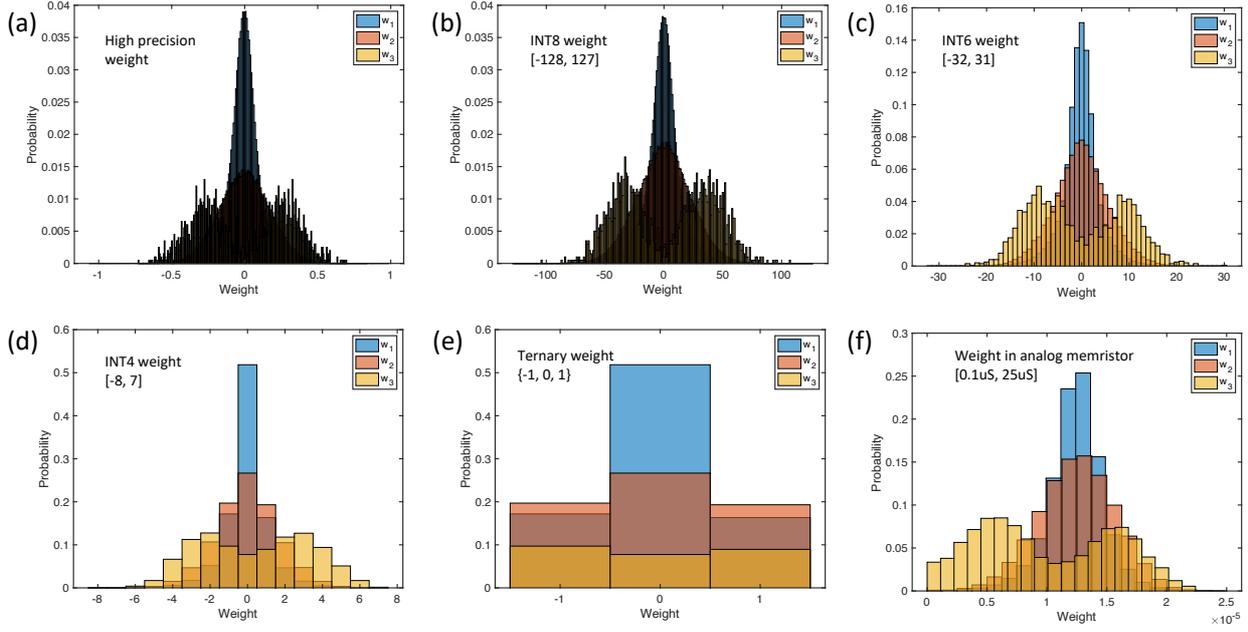

**Extended Data Figure 8 | The learned weight distribution in the quantized and memristor-based neural network compared with the high-precision weight for binary stochastic learning.** **(a)** High-precision weight, **(b)** INT8 weight (scaling factor: 128), **(c)** INT6 weight (scaling factor: 32); **(d)** INT4 weight (scaling factor: 8), **(e)** Ternary weight (scaling factor: 2), **(f)** analog memristor weights ($V_0 = 2V$, $G_0 = 25$ uS, $G_{ref} = 13$ uS). The ranges of the weight distribution are labeled in each panel.

**Extended Data Table II.** The energy consumption of the elementary multiply & accumulation (MAC) operations in various learning schemes.

| Weight type \ Learning type | HP learning | BS learning |
|---|---|---|
| FP32 | FP32*FP32+FP32 (4.6 pJ)[a] | 1bit*FP32+FP32 (0.9 pJ)[a] |
| INT8 | -- | 1bit*INT8+INT8 (0.03 pJ)[a] |
| INT4 | -- | 1bit*INT4+INT4 (~15 fJ) |
| Ternary | -- | 1bit*1.5bit+1.5bit (~5.6 fJ) |
| Analog memristor | 8bit-input, ADC-output (0.18 pJ)[b] | 1bit-input, No-ADC (~1.8 fJ) |

[a] 45nm, 0.9V, CMOS technology [M. Horowitz, ISSCC, 2014];

[b] 65nm, RRAM memristor, 128*128 1T1R array [P. Yao, Nature, 2020].